\documentclass[11pt]{article}

\usepackage[preprint]{acl}

\usepackage{times}
\usepackage{latexsym}

\usepackage[T1]{fontenc}

\usepackage[utf8]{inputenc}

\usepackage{microtype}

\usepackage{inconsolata}

\usepackage{graphicx}

\usepackage{algorithm}

\usepackage[table]{xcolor}%
\usepackage[normalem]{ulem}
\usepackage{multirow}%
\usepackage{amsmath,amssymb,amsfonts}%
\usepackage{amsthm}%
\usepackage{booktabs}%
\usepackage{hyperref}
\usepackage{booktabs,tabularx,array}
\usepackage[most]{tcolorbox}
\usepackage{graphicx}
\usepackage{makecell}
\usepackage{amsmath,algorithmicx,algpseudocode}
\usepackage{xurl}
\usepackage{longtable}

\title{SPIO: Ensemble and Selective Strategies via LLM-Based Multi-Agent Planning in Automated Data Science}

\author{
 \textbf{Wonduk Seo\textsuperscript{1*}}\hspace{0.5em}
 \textbf{Juhyeon Lee\textsuperscript{2}}\thanks{denotes co-first authors.}\hspace{0.5em}
\textbf{Yanjun Shao\textsuperscript{3}}\hspace{0.5em}\\
 \textbf{Qingshan Zhou\textsuperscript{2}}\hspace{0.5em}
 \textbf{Seunghyun Lee\textsuperscript{1}}\thanks{denotes corresponding authors.}\hspace{0.5em}
 \textbf{Yi Bu\textsuperscript{2}}\textsuperscript{\scriptsize †}\hspace{0.5em}
\\
\\
 \textsuperscript{1}Enhans\hspace{1.0em}
 \textsuperscript{2}Peking University\hspace{1.0em}
 \textsuperscript{3}Yale University
\\
\textsuperscript{1}\texttt{\{wonduk,seunghyun\}@enhans.ai}\\
\textsuperscript{2}\texttt{leejuhyeon@stu.pku.edu.cn}, 
\textsuperscript{2}\texttt{\{buyi,zqs\}@pku.edu.cn}\\
\textsuperscript{3}\texttt{yanjun.shao@yale.edu}\\
}

\begin{document}
\maketitle
\begin{abstract}
Large Language Models (LLMs) have enabled dynamic reasoning in automated data analytics, yet recent multi-agent systems remain limited by rigid, single-path workflows that restrict strategic exploration and often lead to suboptimal outcomes. To overcome these limitations, we propose \textbf{SPIO} (\textbf{\underline{S}}equential \textbf{\underline{P}}lan \textbf{\underline{I}}ntegration and \textbf{\underline{O}}ptimization), a framework that replaces rigid workflows with adaptive, multi-path planning across four core modules: data preprocessing, feature engineering, model selection, and hyperparameter tuning. In each module, specialized agents generate diverse candidate strategies, which are cascaded and refined by an optimization agent. \textbf{SPIO} offers two operating modes: \textbf{SPIO-S} for selecting a single optimal pipeline, and \textbf{SPIO-E} for ensembling top-k pipelines to maximize robustness. Extensive evaluations on Kaggle and OpenML benchmarks show that \textbf{SPIO} consistently outperforms state-of-the-art baselines, achieving an average performance gain of 5.6\%. By explicitly exploring and integrating multiple solution paths, \textbf{SPIO} delivers a more flexible, accurate, and reliable foundation for automated data science.

\end{abstract}

\section{Introduction}

Automated data analytics and predictive modeling have emerged as pivotal components in modern data-driven research~\cite{erickson2020autogluon,ebadifard2023data,hollmann2023large,guo2024ds}, empowering practitioners to extract meaningful insights from complex, high-dimensional datasets~\cite{drori2021alphad3m,hollmann2023large,hassan2023chatgpt,yang2024unleashing}. Traditionally, constructing robust analytics pipelines has necessitated extensive manual design and iterative refinement, relying heavily on domain expertise and coding practices with established libraries and frameworks~\citep{bergstra2012random,elsken2017simple,sparks2017keystoneml,de2021automating}. This conventional paradigm, while effective in certain settings, often imposes rigid workflow structures that limit strategic exploration and impede the integration of diverse solution paths, ultimately constraining the scalability and robustness of predictive systems~\citep{sparks2017keystoneml,xin2018helix,maymounkov2018koji,nikitin2022automated}.

\begin{figure}[t]
  \centering
  \includegraphics[width=1\linewidth]{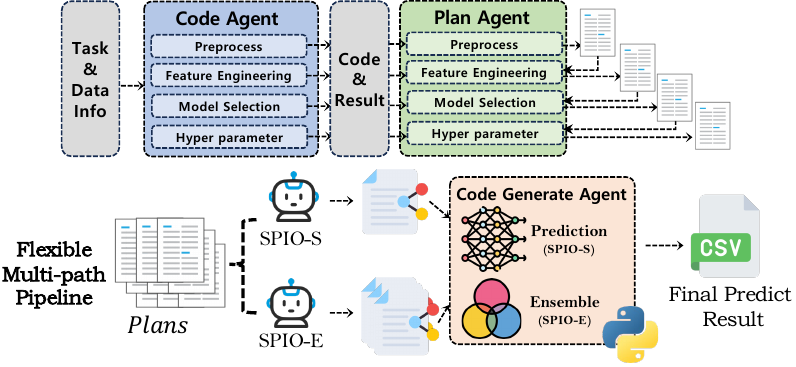}
    \caption{\textbf{SPIO framework overview.} Given dataset descriptions, \textbf{SPIO} produces baseline code and results for each pipeline module. A sequential planning agent then proposes candidate improvements per module, after which \textbf{SPIO} either selects a single best end-to-end path (\textbf{SPIO-S}) or ensembles the top-$k$ paths (\textbf{SPIO-E}) to generate the prediction file.}
  \label{fig:enter-label}
\end{figure}

Large Language Models (LLMs) have become powerful tools for automating data analytics and predictive modeling, addressing prior challenges via natural language instructions~\citep{zhang2023automl,guo2024ds,xu2024large,tayebi2024large,gu2024large}. By translating high-level instructions into executable code, LLM-based systems greatly reduce the need for specialized programming skills and simplify the design of complex analytical workflows~\citep{nijkamp2022codegen,joel2024survey,xu2024aios}. Recent frameworks include Spider2v~\citep{cao2024spider2}, which automates analysis by generating SQL/Python code and manipulating GUI components, and OpenAgents~\citep{xie2023openagents}, a platform that integrates diverse AI agents for unified data analysis and web search~\citep{xie2023openagents}.

More recently, multi-agent systems such as Agent K v1.0~\citep{grosnit2024large}, AutoKaggle~\citep{li2024autokaggle}, and Data Interpreter~\citep{hong2025data} automate end-to-end data analytics. Agent K v1.0 coordinates agents for cleaning, feature engineering, modeling, and hyperparameter tuning; AutoKaggle follows a six-stage pipeline from exploration to planning; and Data Interpreter constructs task graphs that decompose subtasks and dependencies to guide actions. More general-purpose agent frameworks, including OpenHands~\citep{wang2024openhands} and AIDE~\citep{jiang2025aide}, further incorporate execution-driven loops for autonomous code generation and refinement. However, these methods still (1) rely on single-path execution that limits exploration, (2) follow rigid, predefined workflows that struggle with complex tasks, and (3) lack mechanisms to aggregate multi-dimensional feedback, reducing their ability to exploit nuanced signals for optimal predictions.

To address these limitations, we propose \textbf{SPIO} (\underline{\textbf{S}}equential \underline{\textbf{P}}lan \underline{\textbf{I}}ntegration and \underline{\textbf{O}}ptimization), a framework for automated data analytics and predictive modeling. Unlike rigid single-path or fixed multi-agent workflows, \textbf{SPIO} organizes the pipeline into four modules: data preprocessing, feature engineering, model selection, and hyperparameter tuning, and performs sequential planning within each module. At each stage, \textbf{SPIO} records intermediate outputs such as transformed data summaries or validation scores, which interact with results from previous stages. Based on these accumulated plan records, \textbf{SPIO} applies LLM-driven selection and ensembling to optimize end-to-end performance. Code-generation agents first produce executable solutions from a data description \(D\) and a task description \(T\). A sequential planning agent then revises these solutions using data statistics for preprocessing and feature engineering, and validation score for model selection and hyperparameter tuning. Finally, \textbf{SPIO} supports two variants: \textbf{SPIO-S}, which selects a single best pipeline for execution, and \textbf{SPIO-E}, which selects the top \(k\) pipelines and ensembles their predictions.

Extensive experiments on regression and classification benchmark datasets, including Kaggle and OpenML demonstrate \textbf{SPIO}'s superiority over state-of-the-art methods. \textbf{SPIO}'s adaptive multi-path reasoning integrates diverse insights, overcoming the limitations of fixed workflows. Consequently, it improves predictive accuracy, adapts to varied data through iterative feedback, and enhances execution reliability. These strengths make \textbf{SPIO} a valuable solution for business intelligence, scientific research, and automated reporting, where accuracy and flexibility are essential.

\section{Related Work}
\subsection{Machine Learning and AutoML}
Machine learning (ML) develops algorithms that learn from data to make predictions or decisions~\citep{sutton1998reinforcement,bishop2006pattern,das2015applications,helm2020machine}. Despite major advances in ML algorithms~\citep{kirkos2008support,el2015machine,xie2016unsupervised,sarker2021machine}, building effective systems still requires substantial manual effort: practitioners must choose models, craft features, and tune hyperparameters, which is time-consuming and expert-dependent. Automated Machine Learning (AutoML) aims to reduce this burden by automating key steps of pipeline construction~\citep{shen2018automated,wang2021autods,feurer2022auto}. Early systems such as Auto-WEKA~\citep{thornton2013auto} and TPOT~\citep{olson2016tpot} used Bayesian optimization and genetic programming to search predefined model spaces, while later frameworks (e.g., DSM~\citep{kanter2015deep}, OneBM~\citep{lam2017one}, AutoLearn~\citep{kaul2017autolearn}, H2O AutoML~\citep{ledell2020h2o}) incorporated meta-learning and ensembles. However, many AutoML solutions remain constrained by static search spaces, limiting ptation to diverse data and evolving tasks. Thus, scalability and flexibility remain open challenges.


\begin{figure*}[t]
  \centering
  \includegraphics[width=1\linewidth]{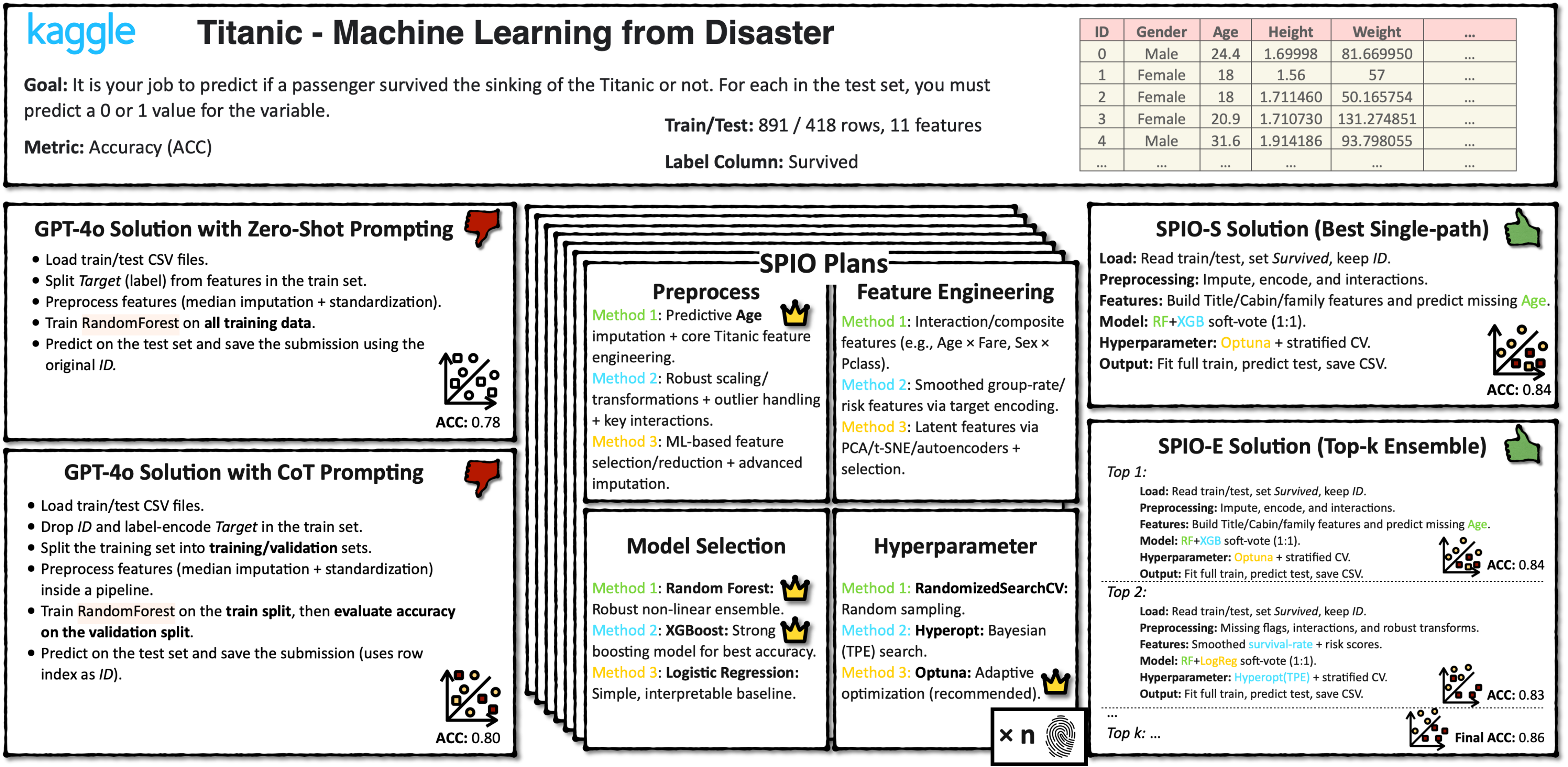}
    \caption{\textbf{Sequential plan integration and optimization in SPIO (Titanic example).} \textbf{SPIO} enumerates candidate plans for preprocessing, feature engineering, model selection, and hyperparameter tuning (e.g., predictive Age imputation, interaction/target-encoded features, RF/XGBoost/LogReg choices, and Optuna/Hyperopt search), then composes them into complete pipelines. Compared with GPT-4o baselines (Zero-Shot ACC 0.78; CoT ACC 0.80), \textbf{SPIO} selects the best single path (\textbf{SPIO-S}, ACC 0.84) or ensembles the top-$k$ pipelines (\textbf{SPIO-E}, final ACC 0.86).}
    \label{fig:ours_framework}
\end{figure*}

\subsection{Data Science Agents}
Large Language Models (LLMs) have accelerated automated data science agents for task understanding, data processing, and predictive modeling. Tools such as LiDA~\citep{dibia2023lida} and GPT4-Analyst~\citep{cheng2023gpt} automate exploratory data analysis, while recent work further leverages code-interpreting capabilities to enable more flexible and complex analytics~\citep{seo2025vispath}. 
Multi-agent systems extend this line toward end-to-end pipelines: 
Agent K~\citep{grosnit2024large} decomposes for cleaning, feature engineering, modeling and tuning. AutoKaggle~\citep{li2024autokaggle} follows a multi-phase workflow, and Data Interpreter~\citep{hong2025data} structures subtasks as dependency graphs; DS Agent Survey~\citep{wang2025large} provides a taxonomy linking agent design to workflow stages, and Blackboard DS Discovery~\citep{salemi2025llm} coordinates specialists via a shared blackboard for scalable data-lake discovery; more general-purpose agent frameworks such as OpenHands~\citep{wang2024openhands} and AIDE~\citep{jiang2025aide} adopt execution-driven iterative loops, refining LLM-generated code through repeated planning, execution, and feedback. Recent benchmarks~\citep{tang2023ml,chan2024mle,huang2023mlagentbench, nathani2025mlgym} have begun to standardize the evaluation of LLM-based data science agents on end-to-end AutoML tasks. However, these systems typically follow a single reasoning path with rigid workflows, which can limit adaptability to task-specific nuances. To address these limitations, we propose the \textbf{SPIO} framework.

\section{Methodology}
\subsection{Preliminaries}
Our proposed \textbf{SPIO} framework, as shown in Figure~\ref{fig:ours_framework}, integrates sequential planning and optimization into an automated analytics pipeline. The framework accepts two primary inputs: a \emph{Data Description} \(D\) (detailing data shape, column types, length, and sample records) and a \emph{Task Description} \(T\) (specifying the prediction type, target column, and relevant background). The framework is organized into four key components, each playing a distinct role in the overall process:

\begin{enumerate}
  \item \textbf{Fundamental Code Generation Agents} generate baseline solutions for each module (preprocessing, feature engineering, model selection, hyperparameter tuning) using \(D\) and \(T\).
  \item \textbf{Sequential Planning Agent} reviews the baseline outputs and based on prior plans, produces up to \(n\) candidate strategies for further improvement.
  \item \textbf{SPIO-S: Single-Path Selection} leverages the candidate plans to select one optimal solution path for final execution.
  \item \textbf{SPIO-E: Ensemble of Sequential Paths} aggregates the top \(k\) candidates via an ensemble approach to enhance predictive robustness.
\end{enumerate}

\subsection{Fundamental Code Generation Agents}

In each module, the fundamental code generation agent establishes a stable baseline by generating executable code and producing a corresponding description of the processed output. We specifically denote each module’s output as a tuple \((C_i, D_i)\) or \((C_i, V_i)\), where \(C_i\) is the generated code, \(D_i\) is the description of the resulting data, and \(V_i\) is the validation score (used in modeling and tuning stages). For instance, given \(D\) and \(T\), the \textit{data preprocessing agent} produces:
\begin{equation}
(C_{\text{pre}}, D_{\text{pre}}) = \mathcal{G}(D, T),
\end{equation}
where \(D_{\text{pre}}\) encapsulates key details for data preprocessing such as data shape, column types, and sample records. The \textit{feature engineering agent} then takes \((C_{\text{pre}}, D_{\text{pre}})\) along with \(T\) to generate:
\begin{equation}
(C_{\text{feat}}, D_{\text{feat}}) = \mathcal{G}(C_{\text{pre}}, D_{\text{pre}}, T).
\end{equation}
For model selection and hyperparameter tuning, the agents output both the generated code and a validation score. Specifically, the \textit{model selection agent} processes \((C_{\text{feat}}, D_{\text{feat}}, T)\) to yield:
\begin{equation}
(C_{\text{model}}, V_{\text{model}}) = \mathcal{G}(C_{\text{feat}}, D_{\text{feat}}, T),
\end{equation}
and the \textit{hyperparameter tuning agent} refines the model further by producing:
\begin{equation}
(C_{\text{hp}}, V_{\text{hp}}) = \mathcal{G}(C_{\text{model}}, D_{\text{feat}}, T).
\end{equation}
These outputs form the foundation for subsequent reflective improvement.

\subsection{Sequential Planning Agent}
After each fundamental code generation step, a sequential planning agent is applied to refine the outputs by proposing alternative strategies. In the data preprocessing and feature engineering modules, the planning agent uses the current code \(C_i\) and its corresponding output description \(D_i\) (for \(i \in \{\text{pre}, \text{feat}\}\)), along with \(D\), \(T\), and all previously generated candidate plans \(P_{<i}\), to propose up to \(n\) alternative strategies:
\begin{equation}
\{P_i^1, P_i^2, \dots, P_i^n\} = \mathcal{G}(C_i, D_i, D, T, P_{<i}).
\end{equation}
For the model selection and hyperparameter tuning modules, the planning agent incorporates the validation score \(V_i\) (with \(i \in \{\text{model}, \text{hp}\}\)) as a key performance indicator, generating:
\begin{equation}
\{P_i^1, P_i^2, \dots, P_i^n\} = \mathcal{G}(C_i, V_i, D, T, P_{<i}).
\end{equation}
This reflective planning mechanism systematically explores alternative solution paths, thereby supporting subsequent optimization.

\subsection{SPIO-S: Single-Path Selection}
The \textbf{SPIO-S} variant employs an LLM-driven optimization agent to directly identify and select the single optimal solution path from among candidate plans generated across all modules. Formally, let
\[
\mathcal{P} = \bigcup_{i \in \{\text{pre},\,\text{feat},\,\text{model},\,\text{hp}\}}\{P_i^1, P_i^2,\dots,P_i^n\}
\]
denote the complete set of candidate strategies. Given the data description \(D\), task description \(T\), and intermediate module outputs, the LLM autonomously evaluates and selects the most effective candidate plan:
\begin{equation}
P^* = \text{SelectBest}_{\text{LLM}}(\mathcal{P}, D, T).
\end{equation}
The final executable code is subsequently generated by applying the generative model to the chosen plan \(P^*\):
\begin{equation}
C_{\text{final}} = \mathcal{G}(P^*, D, T).
\end{equation}
Executing \(C_{\text{final}}\) yields the final predictive outcome.

\subsection{SPIO-E: Ensemble of Sequential Paths}
The \textbf{SPIO-E} variant enhances predictive robustness by utilizing an LLM-driven optimization agent to select the top \(k\) candidate plans for ensemble integration. Specifically, given the candidate plan set \(\mathcal{P}\), along with descriptions \(D\), \(T\), and intermediate module outputs, the LLM autonomously selects and ranks the most promising \(k\) candidates:
\begin{equation}
\{P^*_1, P^*_2,\dots,P^*_k\} = \text{SelectTopK}_{\text{LLM}}(\mathcal{P}, D, T).
\end{equation}
For each selected candidate plan \(P_i^*\), executable code is independently generated:
\begin{equation}
C_{\text{final}}^i = \mathcal{G}(P^*_i, D, T).
\end{equation}
For classification tasks, the ensemble prediction is computed through soft voting:
\begin{equation}
\hat{y} = \arg\max_{c\in\mathcal{C}}\left(\frac{1}{k}\sum_{i=1}^{k}p_i(c)\right),
\end{equation}
while for regression tasks, predictions are averaged across the \(k\) models:
\begin{equation}
\hat{y} = \frac{1}{k}\sum_{i=1}^{k}y_i.
\end{equation}

\section{Experiments}
\subsection{Setup}

\paragraph{Models Used.}
We utilize \texttt{GPT-4o}~\citep{achiam2023gpt},
\texttt{Claude 3.5 Haiku}~\citep{anthropic2024model}\footnote{\url{https://www.anthropic.com/claude/haiku}},and the open-weight \texttt{LLaMA3-8B model}~\citep{dubey2024llama}\footnote{\url{https://ai.meta.com/llama/}} for our generation tasks. Following prior LLM-based data-science agents and code-generation frameworks to balance exploration and execution reliability~\cite{cao2024spider2,guo2024ds,wu2024daco}, we adopt a moderate sampling setup for \textbf{SPIO} to encourage diverse while maintaining stable, executable code: temperature$=0.5$, top\_p$=1.0$, and max\_tokens$=4096$.

\paragraph{Experimental Datasets.} 

We evaluate our approach on two widely used benchmarks: OpenML~\citep{vanschoren2014openml} and Kaggle.\footnote{\url{https://www.kaggle.com/}} From OpenML, we use 4 popular datasets (3 regression, 1 binary classification), partitioned into 70\% train, 10\% validation, and 20\% test due to the absence of predefined splits. From Kaggle, we use 8 datasets (7 classification, 1 regression). Since these provide leaderboard test sets, we split the given training data into 70\% training and 30\% validation data. The validation sets are utilized for planning and optimizing, while the leaderboard test sets are reserved for the main experiments.\footnote{Detailed dataset descriptions are provided in Table~\ref{tab:datasets}.}

\paragraph{Baseline Methods and Implementation Details.} 

We compare our approach with two single-inference and five recently proposed representative multi-agent baselines, specifically:
\begin{enumerate}
\item \textbf{Zero Shot Inference:} The LLM generates code tailored to the task, without intermediate reasoning.
\item \textbf{CoT Inference}~\cite{wei2022chain}: The LLM generates an explicit step-by-step reasoning trace to derive the pipeline, then outputs the final executable code.
\item \textbf{Agent K v1.0}~\cite{grosnit2024large}: uses a two-stage workflow that first prepares the data and submission artifacts, then trains models with hyperparameter tuning and ensembling.
\item \textbf{AutoKaggle}~\cite{li2024autokaggle}: follows a six-phase workflow: task/background understanding, exploratory analysis, data cleaning, feature engineering, modeling, and final submission generation, coordinated by specialized agents.
\item \textbf{OpenHands}~\cite{wang2024openhands}: a general-purpose autonomous agent framework that decomposes user instructions into tool-augmented action sequences. It iteratively plans, executes, and revises code using execution feedback, relying on a unified agent loop rather than task-specific phases.
\item \textbf{Data Interpreter}~\cite{hong2025data}: constructs high-level task graphs and iteratively refines them into executable action graphs, translating abstract intent into step-by-step code actions guided by intermediate results.
\item \textbf{AIDE}~\cite{jiang2025aide}: performs ML engineering as code optimization, using execution scores to search a solution tree. It iteratively selects promising nodes, applies small fixes or improvements, and evaluates new variants.
\end{enumerate}
Compared to the baselines, our framework \textbf{SPIO} utilizes LLM-driven planning ability across four modules (preprocessing, feature engineering, modeling, and hyperparameter tuning), generating up to two candidate plans per module, where \textbf{SPIO-S} selects the single best plan and \textbf{SPIO-E} ensembles the top two plans.

\paragraph{Evaluation Metrics.} 

For each task, we use standard, task-appropriate metrics. \textit{Root Mean Squared Error} (RMSE) is the primary metric for regression tasks, and for computational convenience, we report RMSE even when the official metric is MSE. We additionally evaluate performance using \textit{ROC} metrics when applicable, and use \textit{Accuracy} (ACC) as the primary metric for classification tasks.

\begin{table*}[t]
\caption{\textbf{Performance comparison on 8 Kaggle datasets and 4 OpenML datasets for three LLM backends (GPT-4o, Claude 3.5 Haiku, and LLaMA3-8B).}
\textbf{Bold} indicates the best; \underline{underline} indicates the second-best. Human expert denotes the best public Kaggle leaderboard score achieved by a human expert submission (i.e., the top-ranked result at the time of reporting).
An asterisk (*) denotes a statistically significant improvement for \textbf{SPIO} variants (paired t-test with Holm-Bonferroni correction, $p<0.05$). Results for OpenHands and AIDE are unavailable under \texttt{LLaMA3-8B} due to context-length constraints and the lack of native function-calling support.}
  \centering
  \resizebox{\linewidth}{!}{
\begin{tabular}{lcccccccccccc}
    \toprule
    \makecell[c]{\textbf{Methods}}
    & \makecell[c]{\textbf{Titanic}}
    & \makecell[c]{\textbf{Spaceship}}
    & \makecell[c]{\textbf{Monsters}}
    & \makecell[c]{\textbf{Academic}}
    & \makecell[c]{\textbf{Obesity}}
    & \makecell[c]{\textbf{Kc1}}
    & \makecell[c]{\textbf{Bank}\\\textbf{Churn}}
    & \makecell[c]{\textbf{Plate}\\\textbf{Defect}}
    & \makecell[c]{\textbf{House}\\\textbf{Price}}
    & \makecell[c]{\textbf{Boston}}
    & \makecell[c]{\textbf{Diamond}}
    & \makecell[c]{\textbf{Sat11}} \\
    &
    \makecell[c]{(ACC$\uparrow$)} &
    \makecell[c]{(ACC$\uparrow$)} &
    \makecell[c]{(ACC$\uparrow$)} &
    \makecell[c]{(ACC$\uparrow$)} &
    \makecell[c]{(ACC$\uparrow$)} &
    \makecell[c]{(ACC$\uparrow$)} &
    \makecell[c]{(ROC$\uparrow$)} &
    \makecell[c]{(ROC$\uparrow$)} &
    \makecell[c]{(RMSE$\downarrow$)} &
    \makecell[c]{(RMSE$\downarrow$)} &
    \makecell[c]{(RMSE$\downarrow$)} &
    \makecell[c]{(RMSE$\downarrow$)} \\
    \midrule

    \rowcolor{gray!10}\multicolumn{13}{l}{\textbf{GPT-4o}} \\
    ZeroShot
    & 0.7464 & 0.7706 & 0.7051 & 0.8269 & 0.8864 & 0.8499
    & 0.8192 & 0.8640
    & 0.1509 & 4.6387 & 540.3852 & 1447.1177 \\
    CoT~(\citeyear{wei2022chain})
    & 0.7440 & 0.7718 & 0.7108 & 0.8280 & 0.8859 & 0.8373
    & 0.8719 & 0.8654
    & 0.1447 & 4.6387 & 540.3556 & 1447.1177 \\
    Agent K V1.0~(\citeyear{grosnit2024large})
    & 0.7608 & 0.7810 & 0.6673 & 0.7879 & 0.8356 & 0.8490
    & 0.8796 & 0.8209
    & 0.1437 & 3.6797 & 415.8068 & 1338.4025 \\
    Auto Kaggle~(\citeyear{li2024autokaggle})
    & 0.7781 & 0.7753 & 0.7253 & 0.8275 & 0.8864 & 0.8422
    & 0.8742 & 0.8351
    & 0.1441 & 4.2510 & 417.9387 & 1365.7454 \\
    OpenHands~(\citeyear{wang2024openhands})
    & 0.7528 & 0.7860 & 0.7189 & 0.8164 & 0.8815 & 0.8503
    & 0.8830 & 0.8632
    & 0.1379 & 3.4234 & 569.5314 & 1355.3216 \\
    Data Interpreter~(\citeyear{hong2025data})
    & 0.7679 & 0.7870 & 0.7289 & 0.8297 & 0.8986 & 0.8482
    & 0.8836 & 0.8828
    & 0.1415 & 3.3937 & 420.9826 & 1376.4327 \\
    AIDE~(\citeyear{jiang2025aide})
    & 0.7775 & 0.7949 & 0.7088 & 0.8340 & 0.8973 & 0.8567
    & 0.8877 & \textbf{0.8856}
    & 0.1310 & 3.8445 & 444.0037 & 1319.8435 \\
    SPIO\text{-}S~(Ours)
    & \underline{0.7847}* & \underline{0.8010}* & \underline{0.7316}* & \underline{0.8341}* & \underline{0.9071} & \underline{0.8590}
    & \underline{0.8877}* & 0.8836*
    & \underline{0.1310}* & \underline{3.0312} & \underline{404.6253} & \underline{1290.9073} \\
    SPIO\text{-}E~(Ours)
    & \textbf{0.7871}* & \textbf{0.8034}* & \textbf{0.7410}* & \textbf{0.8359}* & \textbf{0.9072} & \textbf{0.8687}
    & \textbf{0.8885}* & \underline{0.8843}*
    & \textbf{0.1298}* & \textbf{2.9192} & \textbf{398.8893} & \textbf{1268.7817} \\
    \midrule

    \rowcolor{gray!10}\multicolumn{13}{l}{\textbf{Claude 3.5 Haiku}} \\
    ZeroShot
    & 0.7488 & 0.7904 & 0.7089 & 0.8281 & 0.8910 & 0.8594
    & 0.7251 & 0.8657
    & 0.1473 & 4.6387 & 543.1463 & 1446.9613 \\
    CoT~(\citeyear{wei2022chain})
    & 0.7488 & 0.7917 & 0.6994 & 0.8282 & 0.8904 & 0.8547
    & 0.7454 & 0.8685
    & 0.1467 & 4.6421 & 543.1463 & 1440.0644 \\
    Agent K V1.0~(\citeyear{grosnit2024large})
    & 0.7512 & 0.7928 & 0.7127 & 0.8288 & 0.8988 & 0.8639
    & 0.8863 & 0.8813
    & 0.1407 & 2.9710 & 546.9876 & 1445.1732 \\
    Auto Kaggle~(\citeyear{li2024autokaggle})
    & 0.7688 & 0.7884 & 0.7207 & 0.8182 & 0.8933 & 0.7427
    & 0.8459 & 0.8243
    & 0.1480 & 3.2183 & 789.4067 & 1288.3061 \\
    OpenHands~(\citeyear{wang2024openhands})
    & 0.7727 & 0.7830 & 0.7051 & 0.8267 & 0.8894 & 0.8456
    & 0.8656 & 0.8577
    & 0.1445 & 3.1952 & 517.1757 & 1369.5845 \\
    Data Interpreter~(\citeyear{hong2025data})
    & 0.7464 & 0.7940 & 0.6994 & 0.8282 & 0.8951 & 0.8578
    & 0.8744 & 0.8642
    & 0.1384 & \textbf{2.8378} & 565.1463 & 1350.7718 \\
    AIDE~(\citeyear{jiang2025aide})
    & 0.7625 & 0.7893 & \textbf{0.7316} & 0.8335 & 0.8948 & \underline{0.8706}
    & \textbf{0.8887} & \underline{0.8856}
    & 0.1392 & 2.8556 & 560.4698 & \underline{1273.8269} \\
    SPIO\text{-}S~(Ours)
    & \textbf{0.7780}* & \underline{0.7996}* & 0.7278* & \underline{0.8336}* & \underline{0.9058}* & 0.8626
    & 0.8812* & \textbf{0.8867}*
    & \underline{0.1334}* & 2.8418 & \underline{514.3608} & 1279.6327 \\
    SPIO\text{-}E~(Ours)
    & \underline{0.7775}* & \textbf{0.8027}* & \underline{0.7297}* & \textbf{0.8340}* & \textbf{0.9066}* & \textbf{0.8723}
    & \underline{0.8863}* & 0.8830*
    & \textbf{0.1332}* & \underline{2.8409} & \textbf{511.2089} & \textbf{1270.2695} \\
    \midrule

    \rowcolor{gray!10}\multicolumn{13}{l}{\textbf{LLaMA3-8B}} \\
    ZeroShot
    & 0.7410 & 0.7704 & 0.6880 & 0.8148 & 0.8793 & 0.8294
    & 0.7783 & 0.8554
    & 0.1521 & 4.6387 & 547.8507 & 1450.8444 \\
    CoT~(\citeyear{wei2022chain})
    & 0.7434 & 0.7819 & 0.7002 & 0.8181 & 0.8789 & 0.8279
    & 0.7839 & 0.8570
    & 0.1497 & 4.6387 & 546.4380 & 1425.4643 \\
    Agent K V1.0~(\citeyear{grosnit2024large})
    & 0.7512 & 0.7803 & 0.7013 & 0.8052 & 0.8855 & 0.8310
    & 0.8671 & 0.8488
    & 0.1458 & 4.0728 & 544.0589 & 1408.9078 \\
    Auto Kaggle~(\citeyear{li2024autokaggle})
    & 0.7415 & 0.7787 & 0.7018 & 0.8150 & 0.8775 & 0.8293
    & 0.8505 & 0.8501
    & 0.1469 & 4.1962 & 542.8794 & 1372.6835 \\
    Data Interpreter~(\citeyear{hong2025data})
    & 0.7536 & 0.7830 & 0.7022 & 0.8187 & 0.8855 & 0.8362
    & 0.8695 & 0.8613
    & 0.1452 & 4.3823 & 538.0820 & 1378.5213 \\
    SPIO\text{-}S~(Ours)
    & \textbf{0.7583}* & \underline{0.7896}* & \underline{0.7101} & \underline{0.8208}* & \underline{0.8935}* & \underline{0.8424}
    & \underline{0.8733}* & \underline{0.8613}*
    & \underline{0.1398}* & \underline{3.6467} & \underline{533.1050} & \underline{1359.8238} \\
    SPIO\text{-}E~(Ours)
    & \underline{0.7560}* & \textbf{0.7907}* & \textbf{0.7115}* & \textbf{0.8238}* & \textbf{0.8971}* & \textbf{0.8474}
    & \textbf{0.8785}* & \textbf{0.8739}
    & \textbf{0.1388} & \textbf{3.4386} & \textbf{524.1970} & \textbf{1353.8603} \\
    \midrule

    \rowcolor{gray!10}\textbf{Human Expert}
    & - & 0.8218 & 0.8072 & 0.8404 & 0.9116 & -
    & 0.9059 & 0.8898
    & - & - & - & - \\
    \bottomrule
\end{tabular}}%
\label{tab:main_results_all12_accroc_rmse_order}
\end{table*}

\subsection{Experimental Analysis}
\subsubsection{Main Experiment Results}

We compare our proposed framework \textbf{SPIO} with several baseline methods. As shown in Table~\ref{tab:main_results_all12_accroc_rmse_order}, while \textbf{ZeroShot} provides quick initial solutions, it typically yields inconsistent results due to its lack of structured reasoning. \textbf{CoT} prompting slightly improves interpretability by adopting step-by-step reasoning but still remains limited by its single-path exploration. Multi-agent frameworks such as \textbf{Agent K v1.0}, \textbf{Data Interpreter}, and \textbf{AutoKaggle} further enhance performance by decomposing the workflow into specialized stages and coordinating multiple agents; however, their stage schemas are largely predefined and they typically refine within a fixed pipeline template, which can limit systematic exploration of diverse module-level alternatives and their combinations. \textbf{AIDE} and \textbf{OpenHands} further enhance performance through execution-driven improvement loops; however, they rely heavily on interaction with runtime feedback rather than explicit plan reasoning, making results more sensitive to execution budgets and offering weaker plan-level traceability and control.

In contrast, \textbf{SPIO} overcomes the limitations of single-path and rigid multi-agent workflows by integrating sequential optimization with explicit multi-path exploration across data preprocessing, feature engineering, model selection, and hyperparameter tuning. By generating and evaluating multiple candidate plans at each stage, \textbf{SPIO} improves adaptability and reduces the risk of suboptimal local decisions. Crucially, SPIO conditions later-stage decisions on intermediate evidence, rather than optimizing each module in isolation. This staged plan trace also makes the final pipeline more transparent, since SPIO can attribute gains to specific module-level choices.\footnote{Experiments on datasets released after the language model’s pretraining are provided in the Appendix~\ref{app:data_2025}.} In particular, \textbf{SPIO-S} employs an LLM-driven optimization agent to select the most effective end-to-end pipeline, while \textbf{SPIO-E} further enhances robustness by ensembling top-ranked plans to exploit complementary modeling strengths. As a result, \textbf{SPIO} consistently outperforms strong baselines across diverse datasets, achieving an average performance improvement of 5.6\% across 12 benchmark datasets (Table~\ref{tab:main_results_all12_accroc_rmse_order}), which demonstrates its effectiveness in complex, real-world scenarios.

\subsection{Ablation Studies}
To further validate the robustness and effectiveness of \textbf{SPIO}, we conducted comprehensive ablation studies focusing on three key aspects: (1) Performance Comparison of Each Plan, (2) Ensemble Validation, (3) and Module Impact Analysis.

\begin{table*}[t]
\caption{\textbf{Results of Ablation Studies: (1) \textit{Performance Comparison}, (2) \textit{Ensemble Validation} and (3) \textit{Module Impact} on \textbf{SPIO-S} and \textbf{SPIO-E} Variants Across GPT-4o, Claude 3.5 Haiku, and LLaMA3-8B.} \textbf{Bold} text indicates the best performance. \underline{Underlined} text indicates the second-best performance.}
\centering
\small
\setlength{\tabcolsep}{4pt}
\renewcommand{\arraystretch}{1.1}
\begin{tabular}{llcc|cc|cc}
\toprule
\textbf{\raisebox{-2ex}[0pt][0pt]{LLM}}
 & &
\multicolumn{2}{c|}{\textbf{GPT-4o}} &
\multicolumn{2}{c|}{\textbf{Claude 3.5 Haiku}} &
\multicolumn{2}{c}{\textbf{LLaMA3-8B}} \\
\cmidrule(lr){3-4} \cmidrule(lr){5-6} \cmidrule(lr){7-8}
& & \textbf{ACC/ROC $\uparrow$} & \textbf{RMSE $\downarrow$} &
\textbf{ACC/ROC $\uparrow$} & \textbf{RMSE $\downarrow$} &
\textbf{ACC/ROC $\uparrow$} & \textbf{RMSE $\downarrow$} \\
\midrule
\rowcolor{gray!10}\multicolumn{8}{l}{\textit{Performance Comparison}} \\
SPIO-S Top1 & &
\textbf{0.8361} & \textbf{424.6737} &
\textbf{0.8344} & \textbf{449.2421} &
\textbf{0.8187} & \textbf{474.1788} \\
SPIO-S Top2 & &
\underline{0.8308} & \underline{448.2788} &
\underline{0.8339} & \underline{469.6405} &
0.8025 & \underline{480.1354} \\
SPIO-S Top3 & &
0.8248 & 501.7758 &
0.8203 & 504.4984 &
\underline{0.8052} & 496.9982 \\
SPIO-S Top4 & &
0.8193 & 454.9485 &
0.8176 & 596.6416 &
0.7931 & 498.5383 \\
\addlinespace[3pt]
\midrule

\rowcolor{gray!10}\multicolumn{8}{l}{\textit{Ensemble Validation}} \\
SPIO-S & &
0.8361 & \underline{424.6737} &
0.8344 & \underline{449.2421} &
\underline{0.8187} & \underline{474.1788} \\
SPIO-E Ensemble2 & &
\textbf{0.8395} & \textbf{417.6800} &
\underline{0.8365} & \textbf{446.1131} &
\textbf{0.8224} & \textbf{470.4087} \\
SPIO-E Ensemble3 & &
\underline{0.8375} & 427.0149 &
\textbf{0.8476} & 460.0835 &
0.8140 & 486.7630 \\
SPIO-E Ensemble4 & &
0.8341 & 429.5985 &
0.8323 & 524.2414 &
0.8014 & 486.3129 \\
\midrule

\rowcolor{gray!10}\multicolumn{8}{l}{\textit{Module Impact Analysis}} \\
(w/o) Preprocess & & 0.8310 & 434.4996 & 0.8276 & 472.4470 & 0.8331 & 497.3395 \\
(w/o) Feature Eng. & & 0.8274 & 458.6899 & 0.8152 & 483.1569 & 0.8258 & 494.3421 \\
(w/o) Model Select & & 0.8271 & 458.4007 & 0.8272 & 484.9608 & 0.8284 & 503.2921 \\
(w/o) Hyper Param & & 0.8317 & 479.2854 & 0.8230 & 684.8432 & 0.8310 & 526.5389 \\
\addlinespace[3pt]
\bottomrule
\end{tabular}
\label{tab:spio_full_results}
\end{table*}

\subsubsection{Performance Comparison of Each Plans} 

We evaluate the predictive performance of candidate plans ranked by the LLM to verify the efficacy of its selection strategy. We specifically compare the relative ranks and weighted average scores of the top-ranked plans generated at each stage. 

As shown in Table \ref{tab:spio_full_results}, experimental results demonstrated that the first-ranked plan consistently achieved superior performance relative to other lower-ranked alternatives across multiple datasets. This finding validates that the LLM-driven ranking method accurately captures the underlying quality of generated plans, confirming the reliability and rationale behind its decisions. Leveraging the LLM's top-ranked plan selection ensures that \textbf{SPIO} effectively prioritizes the most robust strategies, improving overall predictive performance.

\subsubsection{Ensemble Validation}

We further explore the optimal number of ensemble (k) by analyzing predictive performance across ensembles comprising two to four candidate plans. Table \ref{tab:spio_full_results} summarizes the results across different ensemble sizes of each model. 

The results clearly indicate that ensembles of two candidate plans consistently achieved the highest predictive accuracy, providing an optimal balance of diversity and precision. Ensembles with more than two plans introduced redundancy without significant performance gains, whereas a single-plan approach lacked the robustness afforded by ensemble diversity. We demonstrate that an ensemble consisting of the Top $1$ and Top $2$ plans yields the best performance.
\subsubsection{Module Impact Analysis}
We assess the contribution of each module within \textbf{SPIO} by removing, in turn, the data preprocessing, feature engineering, modeling, and hyperparameter tuning components and comparing performance. 

Table \ref{tab:spio_full_results} results in a distinct contrast in how different components support \textbf{SPIO}’s performance. When feature engineering or hyperparameter tuning is removed, the system’s gains largely disappear, resulting in a pronounced reduction in overall predictive accuracy. Meanwhile, eliminating data preprocessing or the modeling module reduces performance to a lesser extent, though the drop remains clearly observable. This contrast suggests that feature engineering and hyperparameter tuning are the primary drivers behind \textbf{SPIO}’s robust, high-quality results across varied datasets.

\section{Qualitative Study}
\label{sec:qual_eval}
To complement quantitative benchmarks, we conducted an expert qualitative study to assess the practical acceptability of \textbf{SPIO} pipelines. We considered 10 distinct LLM–datasets (e.g., GPT-4o-Titanic) pairs and recruited 10 AI practitioners and graduate-level participants, yielding 100 dataset–participant evaluation pairs. For each dataset, we evaluated six items ranging from the preprocessing outcomes to the optimal methods selected by \textbf{SPIO-S} and \textbf{SPIO-E}, and scored each item using four predefined evaluation criteria. In addition, we included a single forced-choice question to collect expert votes on the most highly rated item among the Top-1 to Top-3 candidates.
\paragraph{Evaluation Dimensions}
All Likert items are grouped into four subjective dimensions:
\begin{itemize}
\item \textbf{Plausibility:} Is the \textbf{SPIO} pipeline reasonable and safe?
\item \textbf{Interpretability:} Are \textbf{SPIO}’s explanations and rationales clearer than typical AutoML tools?
\item \textbf{Diversity and Coverage:} Does the multi-path ensemble explore distinct strategies and cover the space well?
\item \textbf{Usability and Trust:} Would the participant trust and use \textbf{SPIO} in practice or research?

\end{itemize}
\paragraph{Result of Metric-Wise Plan Rating}: Participants rated the metric-wise plans of final \textbf{SPIO-S} and \textbf{SPIO-E} pipelines on four dimensions using five-point Likert items. As shown in Table~\ref{tab:qual_spios_spioe_dims}, All stage-wise components average above 4.0, and final pipelines score higher: \textbf{SPIO-S} = 4.24$\pm$0.18 and \textbf{SPIO-E} = 4.45$\pm$0.15.

\begin{table}[t]
\caption{\textbf{Expert subjective ratings (mean$\pm$std) for the final optimal pipelines.} $^{*}$ indicates \textbf{SPIO-E} significantly outperforms \textbf{SPIO-S} (paired t-test with Holm-Bonferroni correction; $p<0.05$).}
\centering
\small
\setlength{\tabcolsep}{6pt}
\begin{tabular}{lcc}
\toprule
\textbf{Dimension} & \textbf{SPIO-S} & \textbf{SPIO-E} \\
\midrule
Plausibility            & 4.38 $\pm$ 0.56 & 4.54 $\pm$ 0.29$^{*}$ \\
Interpretability        & 4.21 $\pm$ 0.45 & 4.37 $\pm$ 0.30$^{*}$ \\
Diversity \& Coverage   & 4.10 $\pm$ 0.47 & 4.35 $\pm$ 0.30$^{*}$ \\
Usability \& Trust      & 4.28 $\pm$ 0.46 & 4.55 $\pm$ 0.26$^{*}$ \\
\bottomrule
\end{tabular}
\label{tab:qual_spios_spioe_dims}
\end{table}

Additionally, Table~\ref{tab:qual_spios_spioe_dims} figures  out that \textbf{SPIO-E} is consistently higher across all dimensions, with statistically significant gains (paired t-test with Holm-Bonferroni correction; $p<0.05$). Finally, to test alignment between the LLM's internal ranking and expert discrete choices, participants selected the best plan among the anonymized top-3 \textbf{SPIO-E} candidates per dataset: Plan~1 was chosen 72/100 times (72\%), versus 22 and 6 for Plan~2 and plan~3, indicating strong agreement with the top-1 recommendation generated from LLM.\footnote{Detailed qualitative evaluation results are provided in Appendix~\ref{app:qual_eval}.}

\section{Discussion}
 
\begin{figure}[t]
  \centering
  \includegraphics[width=\linewidth]{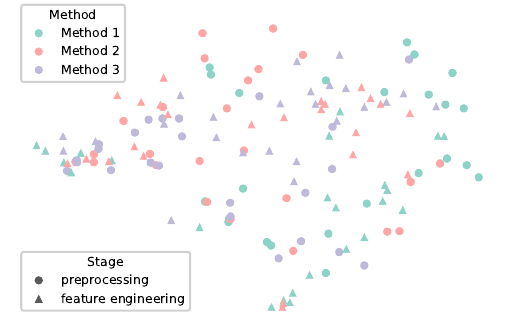}
  \caption{\textbf{Distribution of Preprocessing and Feature Engineering Method.} PCA projection of embedded preprocessing and Feature engineering methods.}
  \label{fig:module_dstrib_preprocess}
\end{figure}

\textbf{Multi-Path Planning improved performance.} Our experiments and ablation studies show that \textbf{SPIO}'s sequential multi-path planning substantially improves automated analytics and predictive modeling, enhancing not only accuracy but also interpretability and robustness. Compared to conventional single-path workflows, \textbf{SPIO} explores multiple candidate strategies at each stage. This systematic yet flexible design produces robust, well-tuned models that better reflect real-world data complexities and user-specific analytical goals. \paragraph{Diverse Planning Across Modules.} To illustrate that \textbf{SPIO} explores diverse strategies rather than following a fixed workflow, we embed the natural-language plans generated in each module using \textit{text-embedding-ada-002} from OpenAI and project them to 2D with PCA for visualization. Figure~\ref{fig:module_dstrib_preprocess} shows that plans in \textit{Preprocessing} and \textit{Feature Engineering} spread broadly in the embedding space, suggesting \textbf{SPIO} adapts its decision to dataset characteristics instead of collapsing to a single preferred pattern.\footnote{Additional module visualizations (model selection and hyperparameter tuning) are provided in Appendix~\ref{app:diversity_vis}.}

\section{Conclusion}
\textbf{SPIO} advances automated data science by integrating sequential planning with multi-path exploration, improving predictive accuracy by an average of 5.6\% over state-of-the-art baselines. \textbf{SPIO} integrates sequential planning, multi-path exploration, and ensemble optimization across four modules, using LLM-driven decisions to generate candidate strategies that capture complex data and improve robustness. Two variants, \textbf{SPIO-S} and \textbf{SPIO-E}, offer flexible use. 
Our ablation studies validate these gains: top-ranked plans outperform lower-ranked alternatives, and a Top-$k=2$ ensemble yields the most reliable improvement, while removing key modules degrades performance. Our discussion highlights that \textbf{SPIO} explores diverse module-level strategies rather than collapsing to a fixed workflow, supporting robustness and interpretability in practice. Future work will extend \textbf{SPIO} to dynamic, interactive data environments to enhance utility in large-scale, real-world deployments.

\section*{Limitation}
Although our baseline comparisons show that \textbf{SPIO} generally achieves stronger predictive performance than single-pass prompting and recent multi-agent frameworks, it does not guarantee that the selected or ensembled pipeline is always globally optimal for every dataset. Since \textbf{SPIO} relies on LLM-generated plans and LLM-driven ranking to compose end-to-end solutions across preprocessing, feature engineering, model selection, and hyperparameter tuning, the final outcome can be sensitive to the LLM backend, prompt context, and generation stochasticity. In future work, we plan to develop more verification- and feedback-driven plan selection, improve efficiency through early pruning and cost-aware search, and expand the supported toolchains to enhance robustness and generalization across broader real-world settings.




\bibliography{custom}

\appendix

\section{Datasets Released After 2025}
\label{app:data_2025}

\begin{table}[h]
\caption{\textbf{Performance on datasets released after LLM pre-training}. 
Results on three Kaggle datasets published in 2025 (RainFall, BackPack, and Podcast), which were unavailable during the pre-training of the evaluated language models. Human expert denotes the best public Kaggle leaderboard score achieved by a human expert submission (i.e., the top-ranked result at the time of reporting). \textbf{Bold} denotes the best result, and \underline{underline} denotes the second-best result. Results for OpenHands and AIDE are unavailable under \texttt{LLaMA3-8B} due to context-length constraints and the lack of native function-calling support.}
  \centering
  \resizebox{\linewidth}{!}{
\begin{tabular}{lccc}
\toprule
\textbf{Methods} &
\textbf{RainFall} &
\textbf{BackPack} &
\textbf{Podcast} \\
& (ACC) $\uparrow$ & (RMSE) $\downarrow$ & (RMSE) $\downarrow$\\
\midrule

\rowcolor{gray!10}\multicolumn{4}{l}{\textbf{GPT-4o}} \\
ZeroShot                & 0.8641 & 38.8250 & 12.3384 \\
CoT                     & 0.8655 & 38.8012 & 12.3158 \\
Agent K V1.0            & 0.8691 & 38.7277 & 12.2849 \\
Auto Kaggle             & 0.8879 & 38.7353 & 12.3042 \\
OpenHands             & 0.8882 & 38.8392 & 12.4421 \\
Data Interpreter        & 0.8884 & 38.7168 & 12.2549 \\
AIDE                    & 0.8932 & 38.8318 & 12.3371 \\

SPIO\text{-}S (Ours)    & \underline{0.8995} & \underline{38.7071} & \underline{12.1873} \\
SPIO\text{-}E (Top 2 Ensemble) & \textbf{0.9004} & \textbf{38.6823} & \textbf{12.1072} \\
\midrule

\rowcolor{gray!10}\multicolumn{4}{l}{\textbf{Claude 3.5 Haiku}} \\
ZeroShot                & 0.8439 & 38.8138 & 12.3419 \\
CoT                     & 0.8497 & 38.8038 & 12.3242 \\
Agent K V1.0            & 0.8526 & 38.7419 & 12.2750 \\
Auto Kaggle             & 0.8719 & 38.7290 & 12.2388 \\
OpenHands             & 0.8789 & 38.9418 & 12.4609 \\
Data Interpreter        & 0.8731 & 38.7249 & 12.1948 \\
AIDE             & 0.8812 & 38.9195 & 12.5523 \\

SPIO\text{-}S (Ours)    & \underline{0.8874} & \underline{38.7071} & \underline{12.1914} \\
SPIO\text{-}E (Top 2 Ensemble) & \textbf{0.8923} & \textbf{38.6672} & \textbf{12.1823} \\
\midrule

\rowcolor{gray!10}\multicolumn{4}{l}{\textbf{LLaMA3-8B}} \\
ZeroShot                & 0.8420 & 38.8450 & 12.3700 \\
CoT                     & 0.8470 & 38.8320 & 12.3550 \\
Agent K V1.0            & 0.8501 & 38.7890 & 12.2980 \\
Auto Kaggle             & 0.8549 & 38.7840 & 12.3090 \\
Data Interpreter        & 0.8564 & 38.7740 & 12.2760 \\
SPIO\text{-}S (Ours)    & \underline{0.8608} & \underline{38.7520} & \underline{12.2450} \\
SPIO\text{-}E (Top 2 Ensemble) & \textbf{0.8738} & \textbf{38.7380} & \textbf{12.2278} \\
\midrule
\rowcolor{gray!10}\textbf{Human Expert} & 0.9065 & 38.6162 & 11.4483\\
\bottomrule
\end{tabular}}
\label{tab:rain_backpack_podcast}
\end{table}
Most datasets used in our main experiments were released prior to 2025. As a result, one might raise concerns that such datasets could have been included in the pre-training corpora of large language models, potentially introducing unintended data leakage. To explicitly address this concern, we additionally evaluate our framework on three Kaggle datasets released in 2025, consisting of one classification task and two regression tasks. These datasets are used solely to verify that the observed performance gains stem from the proposed methodology rather than memorization or prior exposure during model pre-training.

Specifically, we include the RainFall (classification), BackPack (regression), and Podcast (regression) datasets, all of which were published in 2025 and were therefore unavailable during the pre-training of the evaluated language models. Experimental results on these datasets demonstrate that \textbf{SPIO} continues to outperform strong baselines, confirming that the performance improvements are driven by structured multi-path planning and optimization rather than reliance on pre-trained knowledge. Notably, on the RainFall classification dataset, \textbf{SPIO-E} with \texttt{GPT-4o} achieves performance within approximately the \textbf{top 9\%} of the public Kaggle leaderboard at the time of submission, further highlighting the practical competitiveness and robustness of our approach on contemporary, real-world data.

\begin{table*}[t]
\centering
\small
\caption{\textbf{Dataset statistics used in the main experiments.} 
We report dataset sizes, feature dimensionality, task types, and the fraction of missing values in the training and test splits.}
\label{tab:dataset_overview}
\resizebox{\textwidth}{!}{%
\begin{tabular}{l r r r l l c}
\toprule
\textbf{Dataset} & \textbf{\#Train} & \textbf{\#Test} & \textbf{\#Features} & \textbf{Label} & \textbf{Task} & \textbf{Null Ratio (Train/Test)} \\
\midrule
\rowcolor{gray!10}\multicolumn{7}{l}{\textit{Kaggle Datasets}} \\
Titanic             & 891      & 418      & 11  & Survived      & Classification & 0.088 / 0.090 \\
House Prices        & 1460     & 1459     & 80  & SalePrice     & Regression     & 0.067 / 0.068 \\
Spaceship Titanic   & 8693     & 4277     & 13  & Transported   & Classification & 0.021 / 0.020 \\
Monsters            & 371      & 529      & 6   & Type          & Classification & 0.000 / 0.000 \\
Academic Success    & 76518    & 51012    & 37  & Target        & Classification & 0.000 / 0.000 \\
Bank Churn          & 165034   & 110023   & 13  & Exited        & Classification & 0.000 / 0.000 \\
Obesity Risk        & 20758    & 13840    & 17  & NObeyesdad    & Classification & 0.000 / 0.000 \\
Plate Defect        & 19219    & 12814    & 35  & Class         & Multi-class    & 0.000 / 0.000 \\
BackPack            & 3694318  & 200000   & 9   & Price         & Regression     & 0.019 / 0.021 \\
Rainfall            & 2190     & 730      & 11  & Rainfall      & Classification & 0.000 / 0.000 \\
Podcast             & 750000   & 250000   & 10  & Listening\_Time\_minutes & Regression & 0.031 / 0.031 \\
\midrule
\rowcolor{gray!10}\multicolumn{7}{l}{\textit{OpenML Datasets}} \\
Boston Housing      & 506      & --       & 13  & MEDV          & Regression     & 0.000 / -- \\
Diamonds            & 53940    & --       & 9   & Price         & Regression     & 0.000 / -- \\
KC1                 & 2109     & --       & 21  & Defects       & Classification & 0.000 / -- \\
SAT11               & 4440     & --       & 117 & Runtime       & Regression     & 0.000 / -- \\
\bottomrule
\end{tabular}
}
\end{table*}

\section{Supplementary of Main Experiments}
\label{app:supp_result_main}

\paragraph{Dataset Information}
Tables~\ref{tab:dataset_overview} and~\ref{tab:datasets} summarize the datasets used in our main experiments, including their sizes, feature dimensionality, task types, and data sources. 
These benchmarks cover a diverse set of real-world supervised learning tasks from Kaggle and OpenML, spanning binary and multi-class classification as well as regression, with dataset sizes ranging from hundreds to millions of instances. 
\begin{table*}[t]
\centering
\small
\caption{\textbf{Summary of datasets used in our experiments}. The table includes benchmark datasets from Kaggle and OpenML, covering a diverse range of supervised learning tasks.}
\label{tab:datasets}
\begin{tabular}{ll}
\toprule
\textbf{Dataset} & \textbf{Link} \\
\hline
\rowcolor{gray!10}\multicolumn{2}{l}{\textit{Kaggle Datasets}} \\
Titanic &
\url{https://www.kaggle.com/competitions/titanic} \\
House Prices &
\url{https://www.kaggle.com/competitions/house-prices-advanced-regression-techniques} \\
Spaceship Titanic &
\url{https://www.kaggle.com/competitions/spaceship-titanic} \\
Monsters &
\url{https://www.kaggle.com/competitions/ghouls-goblins-and-ghosts-boo} \\
Academic Success &
\url{https://www.kaggle.com/competitions/playground-series-s4e6} \\
Bank Churn &
\url{https://www.kaggle.com/competitions/playground-series-s4e1} \\
Obesity Risk &
\url{https://www.kaggle.com/competitions/playground-series-s4e2} \\
Plate Defect &
\url{https://www.kaggle.com/competitions/playground-series-s4e3} \\
BackPack &
\url{https://www.kaggle.com/competitions/playground-series-s5e2} \\
Rainfall &
\url{https://www.kaggle.com/competitions/playground-series-s5e3} \\
Podcast &
\url{https://www.kaggle.com/competitions/playground-series-s5e4} \\
\hline
\rowcolor{gray!10}\multicolumn{2}{l}{\textit{OpenML Datasets}} \\
Boston Housing &
\url{https://www.openml.org/d/531} \\
Diamonds &
\url{https://www.openml.org/d/42225} \\
KC1 &
\url{https://www.openml.org/d/1067} \\
SAT11 &
\url{https://www.openml.org/d/41980} \\
\bottomrule
\end{tabular}
\end{table*}

\paragraph{Failure Rate Analysis.}
We report the failure rate at $K$ attempts (FR@K) for baseline methods and \textbf{SPIO} variants. FR@K is defined as the proportion of tasks that are \emph{not} successfully completed within at most $K$ execution attempts. For example, FR@3 denotes the fraction of tasks that still fail after three execution attempts.
Table~\ref{tab:fr_gpt_claude} summarizes FR@K for representative baseline methods and \textbf{SPIO-S} using \texttt{GPT-4o} and \texttt{Claude~3.5~Haiku}. Several single-path or weakly validated approaches (e.g., Agent~K~v1.0 and AutoKaggle) exhibit relatively high initial failure rates (FR@1), indicating vulnerability to cascading errors caused by an incorrect first-generation output. More execution-driven methods, such as AIDE and OpenHands, also show non-negligible FR@1, reflecting instability in early iterations despite iterative refinement. In contrast, \textbf{SPIO-S} rapidly reduces failures, achieving FR@3 = 0 under both LLMs despite a non-zero FR@1. This behavior demonstrates effective early recovery enabled by \textbf{SPIO}'s multi-path candidate generation and selection mechanism. All evaluated methods converge to zero failure rates by $K=10$.

\begin{table}[t!]
\caption{\textbf{Failure rate at $K$ attempts (FR@K) across methods using \texttt{GPT-4o} and \texttt{Claude 3.5 Haiku} (lower is better).}}
  \centering
  \small
  \setlength{\tabcolsep}{6pt}
  \renewcommand{\arraystretch}{1.07}
  \begin{tabular}{lcccc}
    \toprule
    \textbf{Score} & \textbf{FR@1} & \textbf{FR@3} & \textbf{FR@5} & \textbf{FR@10} \\
    \midrule
    \rowcolor{gray!10}\multicolumn{5}{l}{\textbf{GPT-4o}} \\
    ZeroShot        & 0.08 & 0    & 0    & 0 \\
    CoT             & 0    & 0    & 0    & 0 \\
    Agent K V1.0    & 0.33 & 0.16 & 0.08 & 0 \\
    Data Interpreter& 0    & 0    & 0    & 0 \\
    Auto Kaggle     & 0.50 & 0.08 & 0.08 & 0 \\
    AIDE            & 0.33 & 0.08 & 0    & 0 \\
    OpenHands       & 0.50 & 0.16 & 0.08    & 0 \\
    SPIO-S (Ours)   & 0.33 & 0    & 0    & 0 \\
    \midrule
    \rowcolor{gray!10}\multicolumn{5}{l}{\textbf{Claude 3.5 Haiku}} \\
    ZeroShot        & 0    & 0    & 0    & 0 \\
    CoT             & 0.25 & 0    & 0    & 0 \\
    Agent K V1.0    & 0.17 & 0.08 & 0    & 0 \\
    Data Interpreter& 0.08 & 0.08 & 0    & 0 \\
    Auto Kaggle     & 0.58 & 0.17 & 0    & 0 \\
    AIDE            & 0.33 & 0.17 & 0    & 0 \\
    OpenHands       & 0.58 & 0.08 & 0    & 0 \\
    SPIO-S (Ours)   & 0.33 & 0.08 & 0    & 0 \\
    \bottomrule
  \end{tabular}
  \label{tab:fr_gpt_claude}
\end{table}

\begin{table}[t!]
  \centering
  \small
  \caption{\textbf{Failure rate of \texttt{LLaMA3-8B} at $K$ attempts of each stage.}}
    \label{tab:fr_modules}
  \setlength{\tabcolsep}{3pt}
  \renewcommand{\arraystretch}{1.07}
  \resizebox{\linewidth}{!}{
  \begin{tabular}{lcccccc}
    \toprule
    \textbf{Step} & \textbf{FR@1} & \textbf{FR@3} & \textbf{FR@5} &
    \textbf{FR@10} & \textbf{FR@15} & \textbf{FR@20} \\
    \midrule
    Feature Engineering & 1.00 & 0.89 & 0.80 & 0.70 & 0.35 & 0 \\
    Preprocess          & 1.00 & 0.84 & 0.75 & 0.56 & 0.25 & 0 \\
    HyperParameter      & 1.00 & 0.80 & 0.69 & 0.47 & 0.25 & 0 \\
    Model Selection     & 1.00 & 0.73 & 0.59 &
                          0.35 & 0.10 & 0 \\
    \bottomrule
  \end{tabular}
  }
\end{table}

To further analyze robustness under constrained model capacity, we measure FR@K at each pipeline stage using \texttt{LLaMA-3~8B} (Table~\ref{tab:fr_modules}). The uniformly high initial failure rates (FR@1 = 1.0 across all stages) reflect the difficulty for a smaller LLM to generate fully executable plans or code on the first attempt without iterative refinement. Nevertheless, failure rates decrease sharply with $K$, often reaching zero by $K=20$ and, in some cases, much earlier. Bold entries denote the lowest FR@K values within each column, corresponding to the earliest recovery among pipeline stages.

\begin{figure}[t]
    \centering
    \includegraphics[width=\linewidth]{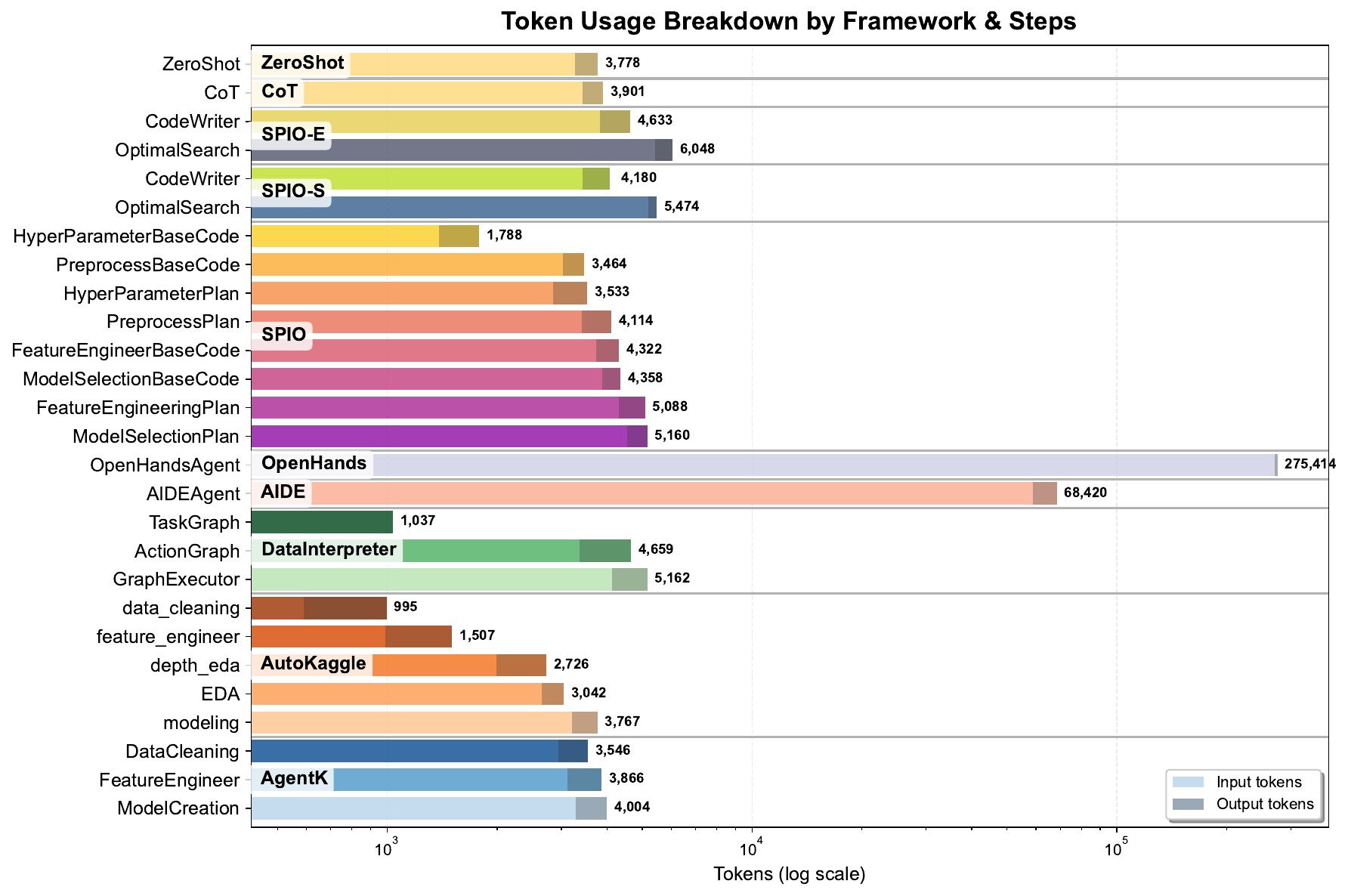}
    \caption{\textbf{Token usage breakdown by framework and steps.} Input and output tokens are shown as stacked bars on a log-scaled x-axis.}
    \label{fig:token_usage_breakdown}
\end{figure}

\paragraph{Token Usage Analysis.}
Figure~\ref{fig:token_usage_breakdown} illustrates the step-wise token usage of all compared frameworks, decomposed into input tokens (prompts, intermediate plans, and execution context) and output tokens (generated code and predictions). Token counts are shown on a logarithmic scale to accommodate the wide variation across methods.

Single-pass baselines such as ZeroShot and CoT exhibit low and stable token usage due to the absence of iterative planning. In contrast, multi-agent pipelines (Agent~K, AutoKaggle, and Data Interpreter) consume substantially more tokens as tasks are decomposed into multiple stages with intermediate representations. OpenHands and AIDE show higher token consumption, with interactions dominated by execution and environment feedback rather than explicit algorithmic planning.

\textbf{SPIO} and its variants maintain bounded per-step token usage while supporting sequential planning and selective exploration across modules. This result indicates that structured, stage-aware multi-path planning can achieve a favorable balance between expressiveness and token efficiency compared to unconstrained autonomous-agent frameworks.

\begin{figure}[t]
    \centering
  \includegraphics[width=\linewidth]{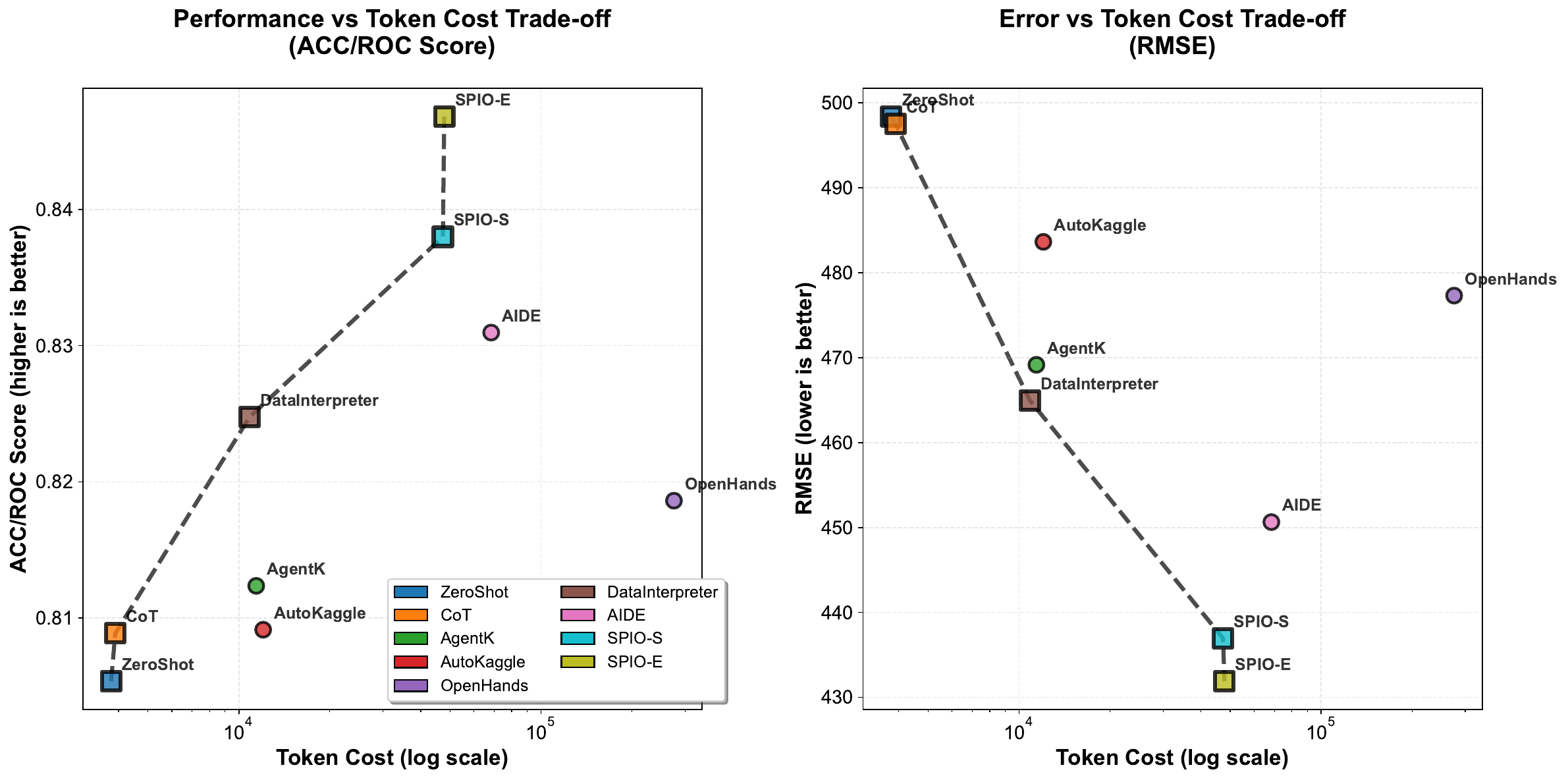}
    \caption{\textbf{Performance versus token cost trade-off across different frameworks.} 
    Left: classification performance measured by ACC/ROC (higher is better). 
    Right: regression performance measured by RMSE (lower is better). 
    Token cost is shown on a logarithmic scale.}
    \label{fig:performance_token_tradeoff}
\end{figure}

\paragraph{Performance--Token Cost Trade-off.}
Figure~\ref{fig:performance_token_tradeoff} analyzes the trade-off between predictive performance and token cost across different frameworks. Each point represents the average performance (ACC/ROC for classification and RMSE for regression) as a function of the corresponding token consumption, plotted on a logarithmic scale. Single-pass baselines achieve low token cost but relatively weaker performance, whereas multi-agent systems improve accuracy at the expense of increased token usage. OpenHands and AIDE incur substantially higher token costs with comparatively limited performance gains. In contrast, \textbf{SPIO-S} and \textbf{SPIO-E} consistently achieve superior performance under comparable or lower token budgets, illustrating that structured multi-path planning yields a more favorable performance--efficiency balance.

\paragraph{Qualitative Planning Analysis.}
We present a concrete comparative example of \textbf{SPIO}’s multi-stage, multi-candidate reasoning on the \textit{Monsters} dataset (\texttt{Claude-3.5 Haiku} variant). This dataset has a moderate size and feature complexity, making it suitable for analyzing the interpretability of alternative solution paths. At each critical stage—preprocessing, feature engineering, model selection, and hyperparameter optimization—\textbf{SPIO} enumerates multiple candidate strategies, evaluates them on validation splits, and retains only the highest-performing pipelines. Table~\ref{tab:monster_paths_compact} summarizes the top four ranked solution paths (Top~1–Top~4) automatically generated and selected.

\begin{table}[t]
\caption{\textbf{Representative high-performing solution paths generated by \textbf{SPIO} on the \textit{Monsters} dataset using Claude~3.5~Haiku}.}
\centering
\footnotesize
\setlength{\tabcolsep}{4pt}
\renewcommand{\arraystretch}{1.1}
\begin{tabularx}{\linewidth}{l c X}
\toprule
\textbf{Rank} & \textbf{Score} & \textbf{Pipeline Configuration} \\
\midrule
Top~1 & 0.7973 &
\textbf{Preprocess:} feature normalization, distribution-aware scaling \newline
\textbf{Feature:} interaction terms, polynomial expansion \newline
\textbf{Model/HPO:} XGBoost + Random Forest ensemble (0.5/0.5), Optuna \\
\addlinespace
Top~2 & 0.6667 &
\textbf{Preprocess:} categorical encoding, feature generation \newline
\textbf{Feature:} probabilistic encoding \newline
\textbf{Model/HPO:} XGBoost + SVM ensemble (0.5/0.5), Bayesian optimization (Hyperopt) \\
\addlinespace
Top~3 & 0.6800 &
\textbf{Preprocess:} robust scaling, outlier handling \newline
\textbf{Feature:} anomaly-aware feature construction \newline
\textbf{Model/HPO:} Random Forest + SVM ensemble (0.5/0.5), grid search (CV) \\
\addlinespace
Top~4 & 0.7255 &
\textbf{Preprocess:} feature normalization \newline
\textbf{Feature:} probabilistic encoding \newline
\textbf{Model/HPO:} XGBoost + Random Forest ensemble (0.5/0.5), Bayesian optimization (Hyperopt) \\
\bottomrule
\end{tabularx}
\label{tab:monster_paths_compact}
\end{table}

\paragraph{Key Distinctions Across Ranked Paths.}
The top-ranked pipelines differ along four principal axes: (i) the depth and rigor of normalization and distribution diagnostics, (ii) the sophistication of categorical and probabilistic encoding strategies, (iii) the exploitation of interaction or polynomial feature expansions, and (iv) the ensemble structure and chosen hyperparameter optimization backend.

\paragraph{Interpretive Factors Driving Top Performance.}
The Top~1 solution outperforms lower-ranked alternatives primarily due to:
\begin{enumerate}
  \item \textbf{Meticulous normalization and outlier handling}, which reduces variance amplification in downstream interaction terms;
  \item \textbf{Explicit modeling of nonlinear feature interactions} via polynomial and cross features, improving class separability;
  \item \textbf{Balanced tree-ensemble composition} (heterogeneous inductive biases of XGBoost and Random Forest), yielding a favorable bias--variance trade-off; and
  \item \textbf{Efficient hyperparameter optimization} (Optuna), producing well-tuned depth, learning rate, and regularization settings within the allocated budget.
\end{enumerate}

Lower-ranked pipelines rely more heavily on categorical embeddings or simpler ensemble combinations (e.g., Random Forest with SVM) and employ less exhaustive interaction generation. As a result, they achieve modestly lower validation accuracy despite remaining methodologically sound. This case study illustrates how \textbf{SPIO}’s evidence-driven ``generate--evaluate--select'' loop at each stage produces robust, high-quality solutions that surpass fixed or less adaptive baselines. The diversity of candidate plans—ranging from normalization-centric to probabilistic encoding and anomaly-aware strategies—also contributes to robustness, as even non-top pipelines remain competitive.

All plans are generated under identical inference settings (temperature 0.5, maximum tokens 4096). For each ranked plan, we log preprocessing transformations, feature engineering steps, selected model families, ensemble weights, and hyperparameter optimization configurations (algorithm, budget, and final parameters). Serialized JSON descriptors are retained for auditability.

\section{Qualitative Expert Evaluation}
\label{app:qual_eval}

\paragraph{Participants and Procedure.}
We recruited 10 participants who either work in AI-related professional roles or hold graduate-level degrees in relevant fields. Each participant evaluated \textbf{SPIO}-generated pipelines on 10 datasets (the same datasets used in the quantitative benchmarks), resulting in a total of 100 dataset--participant evaluation pairs.

For each dataset, \textbf{SPIO} generates plans for the main stages of the data science pipeline—preprocessing, feature engineering, model selection, and hyperparameter tuning—as well as two final ``optimal'' pipelines: one selected by the sequential single-path framework (\textbf{SPIO-S}) and one selected by the ensemble framework (\textbf{SPIO-E}). Participants completed twenty 5-point Likert-scale questions (1 = strongly disagree, 5 = strongly agree) and one multiple-choice question per dataset.

\paragraph{Stage-wise Results.}
Table~\ref{tab:qual_stagewise} reports the average subjective ratings for each \textbf{SPIO} pipeline component and for the two final optimal pipelines, aggregated across all datasets and participants.

\begin{table}[t]
\caption{\textbf{Average subjective ratings (mean and standard deviation) for each \textbf{SPIO} pipeline component and final optimal method (1 = strongly disagree, 5 = strongly agree).}}
\centering
\small
\setlength{\tabcolsep}{6pt}
\begin{tabular}{lcc}
\toprule
\textbf{Pipeline component} & \textbf{Mean} & \textbf{Std. dev.} \\
\midrule
Preprocessing             & 4.07 & 0.17 \\
Feature engineering       & 4.13 & 0.17 \\
Model selection           & 4.20 & 0.17 \\
Hyperparameter selection  & 4.09 & 0.19 \\
\midrule
Optimal method (SPIO-S)   & 4.24 & 0.18 \\
Optimal method (SPIO-E)   & 4.45 & 0.15 \\
\bottomrule
\end{tabular}
\label{tab:qual_stagewise}
\end{table}

\paragraph{Plan-Choice Task (Discrete Agreement with LLM Ranking).}
While Likert-scale ratings capture graded preferences, they do not directly assess whether the LLM’s internal ranking of candidate plans aligns with experts’ discrete choices. To evaluate this alignment, we included a multiple-choice plan-selection task.

\begin{figure}[t]
  \centering
  \includegraphics[width=\linewidth]{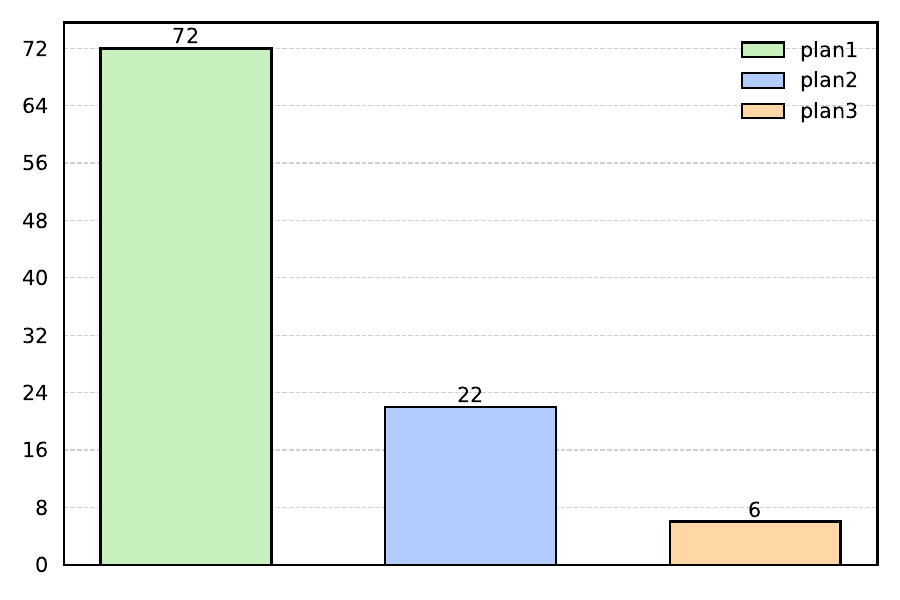}
  \caption{\textbf{Expert choices among the top-3 final plans generated by SPIO-E (Plan~1-3)}. Bars show how often each plan was chosen versus not chosen across all choice tasks.}
  \label{fig:plan_choice_counts}
\end{figure}

For each dataset, \textbf{SPIO-E} generates three candidate ``optimal'' pipelines (Plan~1–Plan~3), corresponding to the top-3 combinations according to the LLM’s internal scoring. As shown in Figure~\ref{fig:plan_choice_counts}, participants were shown these three anonymized pipelines and asked to select the one they considered best suited for the given dataset. This procedure yields 10 datasets $\times$ 10 participants = 100 independent selection tasks. Aggregated results show that Plan~1 was selected in 72 cases, whereas Plan~2 and Plan~3 were chosen 22 and 6 times, respectively.


\begin{figure}[t]
  \centering
  \includegraphics[width=\linewidth]{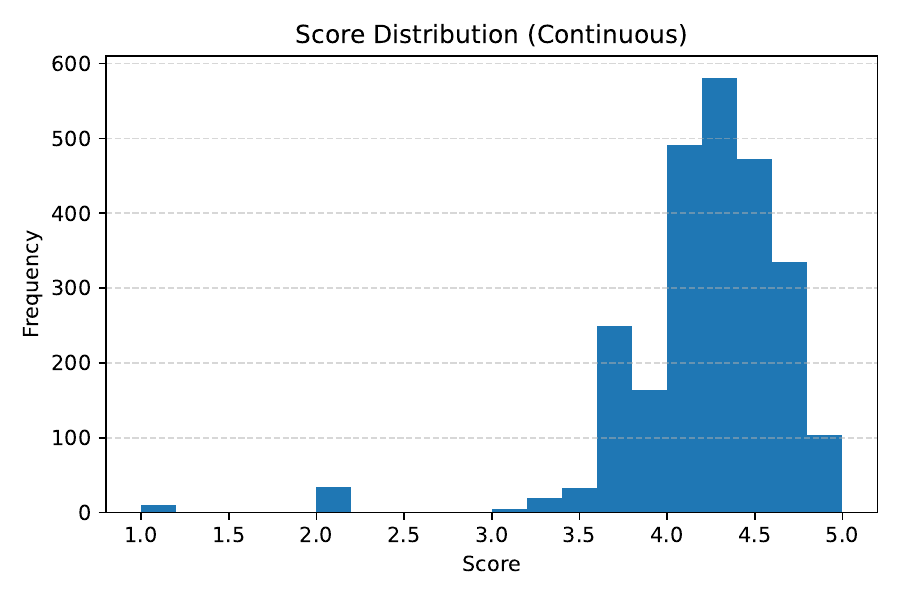}
  \caption{\textbf{Histogram of subjective Likert ratings aggregated over participants, datasets, pipeline stages, and final methods.}}
  \label{fig:qual_score_hist}
\end{figure}

\begin{figure}[t]
  \centering
  \includegraphics[width=\linewidth]{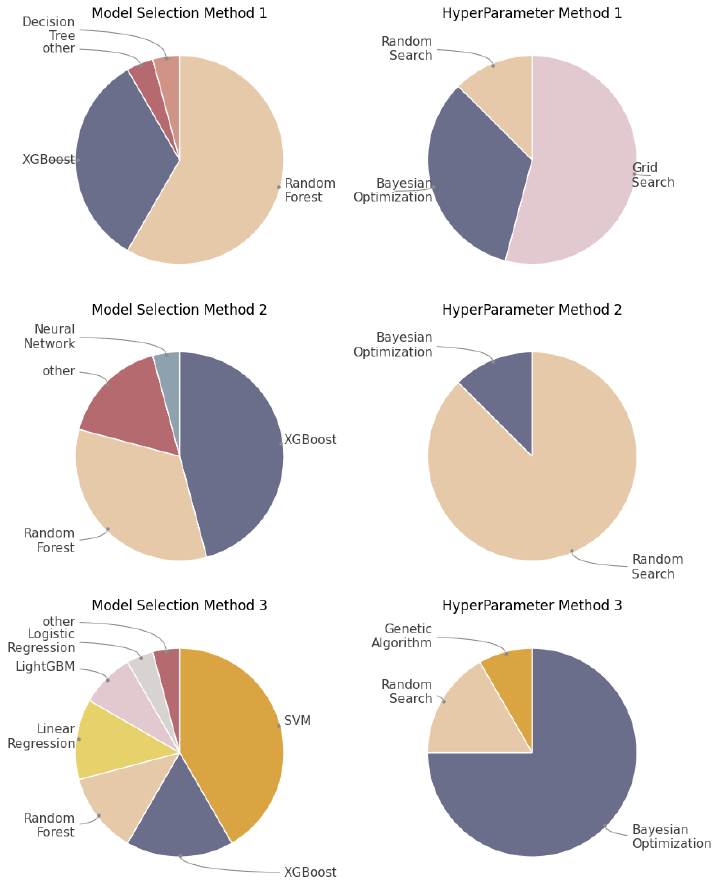}
  \caption{\textbf{Distribution of Model and Hyperparameter selection.} Distribution of methods across the top-3 candidate model-selection and hyperparameter plans.}
  \label{fig:module_dstrib_modelselect}
\end{figure}

\paragraph{Statistical Testing.}
To compare \textbf{SPIO-S} and \textbf{SPIO-E} on the final optimal pipelines, we conducted paired t-tests for each subjective evaluation dimension and applied the Holm--Bonferroni correction to control the family-wise error rate across dimensions. We report statistically significant differences at $p < 0.05$ after correction.

\section{Diversity Visualization Details}
\label{app:diversity_vis}

\paragraph{Embedding and PCA Procedure.}
We visualize the diversity of \textbf{SPIO} generated plans by embedding each plan’s textual description using the \textit{text-embedding-ada-002} model from OpenAI. For a given set of plan texts within a module, we obtain a fixed-dimensional embedding vector for each plan and apply principal component analysis (PCA) to project the embeddings into two dimensions for visualization.

These visualizations are intended as qualitative indicators of diversity. A broad spatial spread and the presence of multiple regions in the two-dimensional projection suggest that the generated plans differ substantially in their semantic content, rather than being minor variations of a single template.

\paragraph{Additional Module Distributions.}
In addition to the preprocessing and feature engineering visualizations presented in the main paper (Figure~\ref{fig:module_dstrib_preprocess}), we also visualize the distributions of plans generated for the \textit{Model Selection} and \textit{Hyperparameter Tuning} modules. Figure~\ref{fig:module_dstrib_modelselect} illustrates how the top candidate plans emphasize different model families and hyperparameter optimization strategies. Although individual plans may prioritize specific design choices, the candidate sets collectively span a wide range of approaches, indicating that \textbf{SPIO} explores diverse alternatives rather than converging on a single preferred tool or configuration.

\paragraph{Interpretation Notes.}
The PCA-based visualizations provide qualitative evidence of plan diversity and should not be interpreted as calibrated or quantitative diversity metrics. In particular, distances in the two-dimensional projection do not necessarily preserve all relationships in the original high-dimensional embedding space, and the embedding model itself induces a specific semantic geometry. Nevertheless, the consistently broad spread observed across modules suggests that \textbf{SPIO} does not collapse to a rigid, single-path pipeline, but instead generates multiple distinct and semantically diverse candidate strategies.

\newpage
\section{Used Libraries in each Module}
\begin{table}[h]
\caption{\textbf{Whitelisted Python libraries/classes by pipeline step}. Only these modules (and Python standard library utilities) are invoked by agent-generated code.}
  \centering
  \scriptsize
  \setlength{\tabcolsep}{6pt}
  \renewcommand{\arraystretch}{1.15}
  \begin{tabularx}{\columnwidth}{>{\raggedright\arraybackslash}p{2cm} >{\raggedright\arraybackslash}X}

    \toprule
    \textbf{Step} & \textbf{Libraries / Classes / Functions} \\
    \midrule
    Preprocessing &
      \texttt{numpy}, \texttt{pandas} \\
    Feature Engineering &
      \texttt{numpy}, \texttt{pandas}, \texttt{sklearn.impute}:\texttt{SimpleImputer},
      \texttt{sklearn.preprocessing}:\texttt{LabelEncoder}, \texttt{MinMaxScaler}, \texttt{PolynomialFeatures}, \texttt{RobustScaler}, \texttt{StandardScaler},
      \texttt{warnings} \\
    Model Selection &
      \texttt{numpy}, \texttt{pandas},
      \texttt{sklearn.compose}:\texttt{ColumnTransformer},
      \texttt{sklearn.ensemble}:\texttt{GradientBoostingRegressor}, \texttt{RandomForestClassifier}, \texttt{RandomForestRegressor},
      \texttt{sklearn.impute}:\texttt{SimpleImputer},
      \texttt{sklearn.linear\_model}:\texttt{Lasso}, \texttt{LinearRegression}, \texttt{LogisticRegression}, \texttt{Ridge},
      \texttt{sklearn.metrics}:\texttt{accuracy\_score}, \texttt{classification\_report}, \texttt{mean\_absolute\_error}, \texttt{mean\_squared\_error}, \texttt{r2\_score}, \texttt{roc\_auc\_score},
      \texttt{sklearn.model\_selection}:\texttt{train\_test\_split},
      \texttt{sklearn.multioutput}:\texttt{MultiOutputClassifier},
      \texttt{sklearn.pipeline}:\texttt{Pipeline},
      \texttt{sklearn.preprocessing}:\texttt{LabelEncoder}, \texttt{OneHotEncoder}, \texttt{StandardScaler} \\
    Hyperparameter Tuning &
      All of the above +
      \texttt{sklearn.model\_selection}:\texttt{GridSearchCV},
      \texttt{sklearn.svm}:\texttt{SVC},
      \texttt{xgboost}:\texttt{XGBClassifier} \\
    \bottomrule
  \end{tabularx}
  \label{tab:used_libs}
\end{table}

\onecolumn
\clearpage
\section{Prompt}
\renewcommand{\arraystretch}{1.3}
\setlength{\tabcolsep}{6pt}

\begin{longtable}{p{0.95\linewidth}}
\caption{\textbf{Prompt for Planning Agent.}}
\label{tab:prompt_planning_agent}
\\
\hline
\textbf{Prompt Template} \\
\hline
\endfirsthead
\hline
\textbf{Prompt Template} \\
\hline
\endhead

Your task:
\begin{itemize}
  \item Your task is to perform: \texttt{\{task\}}.
  \item Focus solely on \texttt{\{task\}}. Performing any unrelated task is strictly prohibited.
  \item Generate a simple and executable code for \texttt{\{task\}}, including explanations of the data and task.
  \item \textbf{Important:} Never drop any rows with missing values in the test data. The length of the test set must remain unchanged.
  \item Your final output should save the preprocessed data as CSV files.
\end{itemize}

\textbf{Input files (read using \texttt{read\_csv}):}
\begin{itemize}
  \item Train data file path: \texttt{\{train\_input\_path\}}
  \item Test data file path: \texttt{\{test\_input\_path\}}
\end{itemize}

\textbf{Output files (save using \texttt{to\_csv}) --- only for preprocessing or feature engineering:}
\begin{itemize}
  \item Train data output path: \texttt{\{train\_output\_path\}}
  \item Test data output path: \texttt{\{test\_output\_path\}}
\end{itemize}

\textbf{Summarized Original Data:} \texttt{\{data\}}

\textbf{Task:} \texttt{\{task\}}

\textbf{Previous Code (excluding preprocessing):} \texttt{\{code\}}
\\
\hline
\end{longtable}


\renewcommand{\arraystretch}{1.3}
\setlength{\tabcolsep}{6pt}

\begin{longtable}{p{0.95\linewidth}}
\caption{\textbf{Prompt for Code Agent.}}
\label{tab:prompt_code_agent}
\\
\hline
\textbf{Prompt Template} \\
\hline
\endfirsthead
\hline
\textbf{Prompt Template} \\
\hline
\endhead

Your task:
\begin{itemize}
  \item You are an expert in \texttt{\{task\}}. Your goal is to solve the task using the given information.
  \item Propose up to \textbf{two alternative methods} (plans) to achieve the task.
  \item For each method, clearly explain:
  \begin{itemize}
    \item In which scenario it is most suitable
    \item Why it is recommended
  \end{itemize}
  \item Identify and recommend the best solution based on the provided context.
\end{itemize}

\textbf{Preprocessing Code:} \texttt{\{pp\_code\}}

\textbf{Summarized Data:} \texttt{\{pp\_data\_result\}}

\textbf{Task Description:} \texttt{\{task\}}
\\
\hline
\end{longtable}


\renewcommand{\arraystretch}{1.2}
\setlength{\tabcolsep}{6pt}

\begin{longtable}{p{0.95\linewidth}}
\caption{\textbf{Prompt for \textbf{SPIO-S} Optimal Method Agent.}}
\label{tab:prompt_spio_s_optimal_agent}
\\ \hline
\textbf{Prompt Template} \\
\hline
\endfirsthead
\hline
\textbf{Prompt Template} \\
\hline
\endhead

\textbf{Your task:}
Select one best plan from each of the following: preprocessing, feature engineering, model selection, and hyperparameter tuning.
Combine these into a single pipeline aimed at achieving the highest validation score.
For model selection, you may use multiple models in an ensemble --- each model should have equal weighting.
Your goal is to determine the best-performing pipeline configuration.
\\[0.35em]

\textbf{Data Summary:} \texttt{\{data\}}\\
\textbf{Task Description:} \texttt{\{task\}}\\
\textbf{Preprocessing Plans:} \texttt{\{preprocess\_plan\}}\\
\textbf{Feature Engineering Plans:} \texttt{\{feature\_engineer\_plan\}}\\
\textbf{Model Selection Plans:} \texttt{\{model\_select\_plan\}}\\
\textbf{Hyperparameter Tuning Plans:} \texttt{\{hyper\_opt\_plan\}}
\\ \hline
\end{longtable}

\renewcommand{\arraystretch}{1.2}
\setlength{\tabcolsep}{6pt}

\begin{longtable}{p{0.95\linewidth}}
\caption{\textbf{Prompt for SPIO-E Optimal Method Agent.}}
\label{tab:prompt_spio_e_optimal_agent}
\\ \hline
\textbf{Prompt Template} \\
\hline
\endfirsthead
\hline
\textbf{Prompt Template} \\
\hline
\endhead

\textbf{Your task:}
Select the top \textbf{two} pipelines by combining one plan from each module: preprocessing, feature engineering, model selection, and hyperparameter tuning. 
For model selection, you may include ensembles --- each model must be equally weighted.
Return the result strictly in \textbf{JSONL format} (a list of JSON objects), without any additional explanation or text.
Each value must exactly match the corresponding original content --- no summarization, simplification, or paraphrasing is allowed.
All fields (\texttt{"preprocess"}, \texttt{"feature\_engineering"}, \texttt{"model\_selection"}, \texttt{"optimal\_hyper\_tool"}) must be clearly and precisely filled.
\\[0.35em]

\textbf{Data Summary:} \texttt{\{data\}}\\
\textbf{Task Description:} \texttt{\{task\}}\\
\textbf{Preprocessing Plans:} \texttt{\{preprocess\_plan\}}\\
\textbf{Feature Engineering Plans:} \texttt{\{feature\_engineer\_plan\}}\\
\textbf{Model Selection Plans:} \texttt{\{model\_select\_plan\}}\\
\textbf{Hyperparameter Tuning Plans:} \texttt{\{hyper\_opt\_plan\}}\\[0.35em]

\textbf{Example Output Format:}
\begin{verbatim}
[
  {
    "rank": "top1",
    "best_combine": {
      "preprocess": "...",
      "feature_engineering": "...",
      "model_selection": "...",
      "optimal_hyper_tool": "..."
    }
  },
  ...
]
\end{verbatim}

Ensure that the output is valid JSON that can be parsed using \texttt{json.load()}.
\\ \hline
\end{longtable}

\clearpage
\section{Case Study}
\begin{figure}[h]
  \centering
  \includegraphics[width=0.8\linewidth]{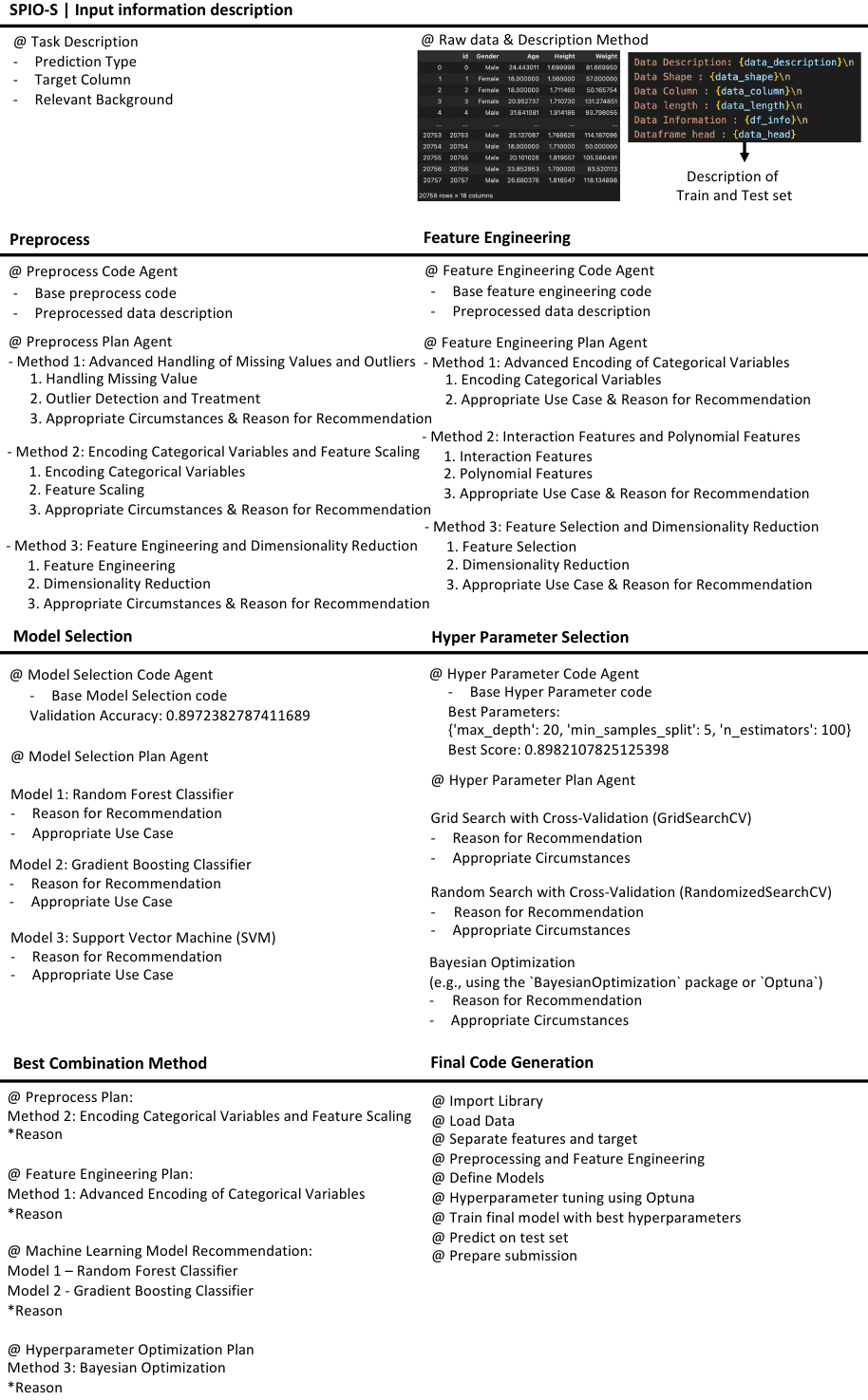}
  \caption{\textbf{An overview of the SPIO-S pipeline, illustrating how agents perform preprocessing, feature engineering, model selection, and hyperparameter tuning}. Among all possible strategies, the system identifies a single best combination plan, which is then used to generate the final executable code.}
  \label{fig:case_study_spios}
  \vspace{-3cm}
\end{figure}

\begin{figure}[h]
  \centering
  \includegraphics[width=0.8\linewidth]{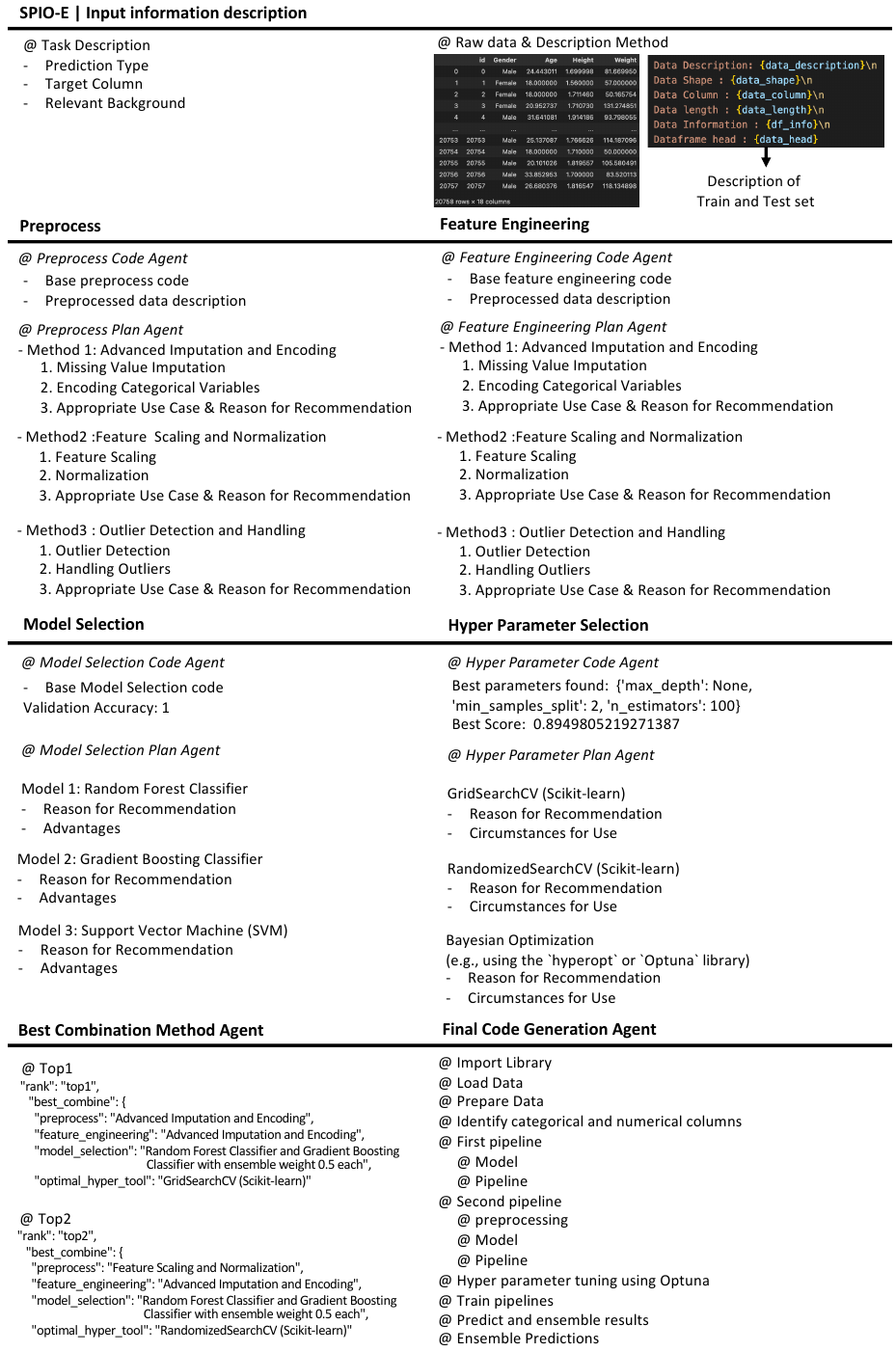}
  \caption{\textbf{An overview of the SPIO-E pipeline, illustrating how agents perform preprocessing, feature engineering, model selection, and hyperparameter tuning}. Based on their outputs, the system selects the 2 best combination plans and produces the final code for execution.}
  \label{fig:case_study_spioe}
\end{figure}

\twocolumn

\end{document}